\documentclass{article}
\usepackage{ijcai26}

\usepackage{times}
\usepackage{soul}
\usepackage{url}
\usepackage[hidelinks]{hyperref}
\usepackage[utf8]{inputenc}
\usepackage[small]{caption}
\usepackage{graphicx}
\usepackage{amsmath}
\usepackage{amsthm}
\usepackage{booktabs}
\usepackage{algorithm}
\usepackage{algorithmic}
\usepackage[switch]{lineno}

\usepackage{helvet}  

\usepackage{graphicx} 
\usepackage{caption} 
\usepackage{amsfonts}
\usepackage{pifont}
\usepackage{amsmath}
\usepackage{xcolor}

\usepackage{booktabs}
\usepackage{multirow}
\usepackage{makecell}
\usepackage{newfloat}
\usepackage{listings}
\DeclareCaptionStyle{ruled}{labelfont=normalfont,labelsep=colon,strut=off} 
\floatstyle{ruled}
\newfloat{listing}{tb}{lst}{}
\floatname{listing}{Listing}
\newcommand{\E}{\mathbb{E}}

\newtheorem{lem}{Lemma}[section]
\newtheorem{thm}{Theorem}[section]

\usepackage[utf8]{inputenc}

\usepackage{url}         
\usepackage{booktabs}  
\usepackage{amsfonts}      
\usepackage{nicefrac}       
\usepackage{microtype}      
\usepackage{xcolor}     
\usepackage{amsthm}

\usepackage{amsmath}
\usepackage{amsfonts}
\usepackage{amssymb}
\usepackage{amsthm}
\usepackage{cuted}
\usepackage{enumitem}
\usepackage{url}

\title{In-Context Reinforcement Learning via Communicative World Models\thanks{Code available at \url{https://github.com/fernando-ml/CORAL}}}

\author{
Fernando Martinez$^1$
\and
Tao Li$^2$\thanks{Corresponding author: li.tao@cityu.edu.hk}\and
Yingdong Lu$^3$\And
Juntao Chen$^1$\\
\affiliations
$^1$Department of Computer and Information Sciences, Fordham University\\
$^2$Department of Systems Engineering, City University of Hong Kong\\
$^3$IBM Research\\
\emails
\{fmartinezlopez, 	jchen504\}@fordham.edu,
li.tao@cityu.edu.hk,
yingdong@us.ibm.com
}

\newif\ifwithappendix
\withappendixtrue 
\begin{document}

\maketitle

\begin{abstract}
Reinforcement learning (RL) agents often struggle to generalize to new tasks and contexts without updating their parameters, mainly because their learned representations and policies are overfit to the specifics of their training environments. To boost agents' in-context RL (ICRL) ability, this work formulates ICRL as a two-agent emergent communication problem and introduces CORAL (Communicative Representation for Adaptive RL), a framework that learns a transferable communicative context by functionally separating latent representation learning from control. In CORAL, an Information Agent (IA) is pre-trained as a world model on a diverse distribution of tasks. 
Its objective is not direct return maximization, but world modeling and
distilling its understanding into concise messages. The emergent communication protocol is shaped by a novel Causal Influence Loss, which measures the effect that the message has on the next action. During deployment, the previously trained IA serves as a fixed contextualizer for a new Control Agent (CA), which learns to solve tasks by interpreting the provided communicative context. Our experiments demonstrate that this approach enables the CA to achieve significant gains in sample efficiency and successfully perform zero-shot adaptation with the help of pre-trained IA in diverse online and offline environments, validating the efficacy of learning a transferable communicative representation. 
\end{abstract}

\section{Introduction}
\label{sec:intro}

Pursuing a generalist agent, which is capable of solving a wide range of tasks with minimal intervention, has long been considered one of the central challenges of reinforcement learning (RL) and artificial intelligence (AI) at large \cite{reed22gato}. Most recently, substantial effort has been dedicated to two distinct research thrusts targeting RL generalization: in-context reinforcement learning (ICRL) \cite{laskin2023incontext} and world models (WM) \cite{silver20wm,hafner25worldmodel}. The key idea behind ICRL is to condition the RL policy on some context variables in addition to the state observations. The adaptation power originates from the agent's response to the emerging context extracted from the past interactions \cite{laskin2023incontext,tao23sce}, which reveals task-related information that improves generalization \cite{tao23cola}. The defining characteristic of ICRL is that the in-context policy improvement is purely context-driven and does not rely on policy model updates as in gradient-based meta RL \cite{finn17maml,tao2024meta,pan-tao25meta-lqr}.

However, most recent ICRL approaches condition on past trajectories using the transformer architecture \cite{vaswani2017attention} but typically do not understand the underlying task dynamics. Consequently, their performance depends largely on the quality of offline datasets \cite{chen2021dt}, which limits generalization beyond the distribution of training contexts \cite{chen2024deep_survery}. In contrast, world models (WMs), typically generative models, aim to equip agents with a structured understanding of the environment dynamics and reward feedback, which allows agents to predict how the environment will evolve in response to their actions, even those rarely seen in the pre-training dataset  \cite{hafnerdream,hafner25worldmodel}. 

From an ICRL perspective, the latent representation learned by the WM, which captures the dynamics and task reward, provides a context for RL agents when deployed in a variety of environments. Our intuition is that WMs are more suitable as contextualizers than vanilla, self-supervised trained transformers, leading to a subsequent reinforcement pre-training in ICRL with improved generalizability \cite{moeini2025survey}. In summary, while ICRL facilitates context-driven adaptation, it often lacks understanding of the environment dynamics, making the generalization fragile. While WMs learn such dynamics, the learned representations, however, are often entangled with task-specific policy learning, which can lead to representations that are overly specialized and less transferable, and the objective mismatch between representation and policy learning \cite{eysenbach2022joint}.

\begin{figure*}[th]
    \centering
    \includegraphics[width=0.875\linewidth]{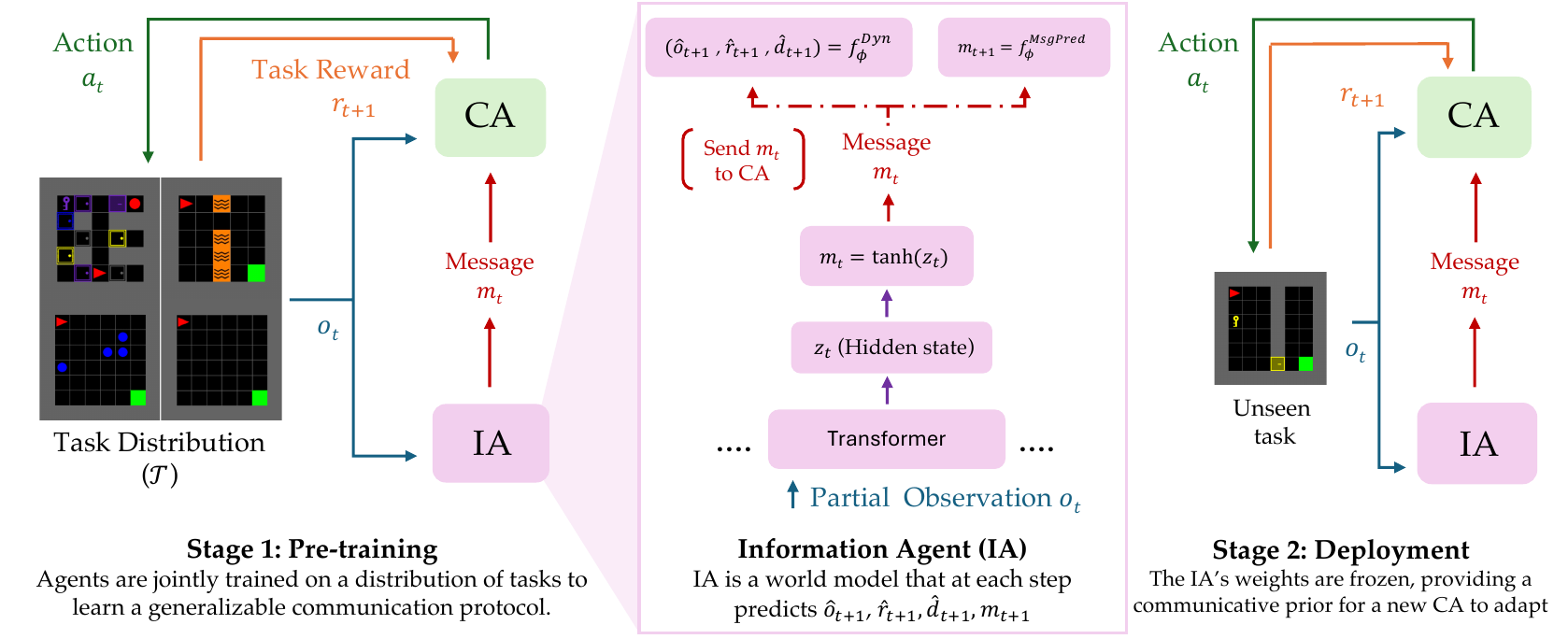}
    \caption{Overview of CORAL. \textbf{(Left) Pre-training:} An Information Agent (IA, $\phi$) and Control Agent (CA, $\theta$) are jointly trained on a task distribution $\mathcal{T}$. The IA processes observations $o_t$ into messages $m_t$ via self-supervised and communication objectives; the CA uses $m_t$ to select actions. \textbf{(Right) Deployment:} The IA is frozen and provides a communicative prior to accelerate a new CA's adaptation to unseen tasks.}
    \label{fig:CORAL-design}
\end{figure*}

To address WMs' limitation above and make it better suited for ICRL, we propose to separate the representation learning from the policy learning and introduce the Communicative Representation for Adaptive RL (CORAL) framework, illustrated in Fig.~\ref{fig:CORAL-design}. CORAL functionally separates the role of world modeling from reward maximization and formulates ICRL as a two-agent emergent communication problem \cite{lowe2019pitfalls,zhu2024survey}, where a world model, called Information Agent (IA), aims to understand the environment dynamics and task reward, and communicate its understanding through latent representations to a Control Agent (CA), parameterized by a neural network policy model. CORAL's novelty lies in that IA, parameterized by a transformer, generates its messages considering three aspects: 1) dynamics awareness, messages encode the predictions of future consequences, 2) temporal coherence, consecutive messages in the same context must be alike to ensure consistency, and 3) in-context communication effectiveness, the control policy conditioned on messages yields higher return than without. 

Our contributions are summarized below, and we postpone the discussion on related works to the end. 1) \textbf{Formulation}: We model a single-agent ICRL problem as a two-agent emergent communication where the world model (information agent) learns how to communicate the learned dynamics to the control agent. 2) \textbf{Hybrid Training}: since the information and control agents face distinct objectives, they are pre-trained differently. The transformer-based IA adopts a self-supervised training with three heads using three losses targeting dynamic awareness, temporal coherence, and communication effectiveness, while the CA adopts proximal policy optimization for reinforcement pre-training. 3) \textbf{Empirical Validation}: We provide extensive empirical validation in partially observable sparse-reward environments and continuous tasks, where CORAL improves sample efficiency most clearly in harder sparse-reward tasks, facilitates zero-shot generalization, and reaches higher performance.

\section{Preliminaries}
We formulate RL tasks as Partially Observable Markov Decision Processes (POMDP)~\cite{kaelbling1998planning}. The agent learns a policy $\pi$ to select actions $a_t$ based on a sequence of partial observations $o_t$, as the true environment state $s_t$ is not directly accessible. The agent's objective is to maximize the expected discounted return $\E_{\pi}[\sum_{t=1}^\infty\gamma^{t-1} r_{t}]$, where $r_t$ is the reward received after taking $a_t$ and $\gamma$ is the discount factor. We define the value function, parameterized by a neural network, as  $V_\theta(o_t) =\E_{\pi}[\sum_{k=0}^\infty \gamma^{k}r_{t+k}|o_t ]$. 

The emergent communication between two agents in a POMDP under a cheap-talk setting is, mathematically, a communication game \cite{crawford82signaling,tao23pot}. In addition to the control agent whose setup is the same as above, an information agent adopts a communication policy $\pi^{\textsc{ia}}(\cdot|o_t)$, from which a message $m_t$ is drawn. The communication is cheap because $m_t$ does not incur cost to any agent nor influence state transition. In this case, the control agent's policy become $\pi^{\textsc{ca}}(\cdot|o_t, m_t)$, which still aims to maximizes the cumulative rewards, whereas the IA's return, fully determined by CA's actions, may share the same objective as CA's \cite{sukhbaatar16comm} or a totally different one \cite{crawford82signaling}.  

\section{Methodology}

The ability to learn representations that generalize well and adapt quickly to new scenarios remains one of the main goals in reinforcement learning. End-to-end solutions in which the loss backpropagated through a single reward signal trains a single monolithic network can produce impressive results~\cite{mnihHumanlevelControlDeep2015}. However, they often produce solutions that are heavily overfitted to the training task and struggle to generalize. Several methods have addressed this problem by decoupling the learning process: allowing agents to learn general-purpose, reusable representations based on self-supervised objectives~\cite{pmlr-v139-stooke21a,schwarzer2021pretraining}. These can learn strong features but may not always align those features with the downstream control task.

\paragraph{Overview.} We introduce CORAL (Communicative Representation for Adaptive RL), a framework that presents a principled middle ground. CORAL frames the problem as one of functional decomposition between representation and control. An Information Agent (IA) learns a dynamics model, while separately, a Control Agent (CA) maximizes reward. The learned protocol by IA can then be flexibly used as a strong contextual prior by CA to enable fast in-context adaptation.

CORAL builds on the asymmetric learning goals of the agents. The CA is a standard RL agent that is motivated by extrinsic task reward. In contrast, 
the IA is not trained on any direct return-maximization objective, and no policy gradient flows from the CA into the IA.
Rather, the IA must learn a predictive model of the world's dynamics and compress this into a communicative representation. To ensure this separated representation remains relevant, the information transmitted by the IA is aligned with the CA's control policy through an influence-theoretic objective.

This alignment is fostered through a composite objective consisting of a self-supervised dynamics modeling loss and a causal influence objective. Constraining the IA to predict future observations, rewards, and terminations causes its internal state to represent physical truths about the environment. Meanwhile, its Causal Influence Loss incentivizes it to send specific messages. In particular, this loss causes the IA to send messages that result in a meaningful change to the CA's policy, but only when that change correlates with high-utility actions. Utility is calculated as a mixture of long-term Generalized Advantage Estimate rewards and the immediate change in the CA's own value prediction, forcing it to learn to communicate both immediately relevant and forward-looking information. Because these two properties are learned from functionally distinct principles, the resultant general representation serves as a powerful pre-trained communicative prior upon deployment.

\subsection{CORAL Agents and Emergent Communication}

CORAL is instantiated with two separate neural networks, which interact with one another at each timestep $t$. Each agent observes the same partial observation $o_t \in \mathbb{R}^{D_{\text{obs}}}$, but has distinct responsibilities. The Information Agent (IA), with parameters $\phi$, aims to store past information and model the environment. We implement this agent with a Transformer-based architecture~\cite{vaswani2017attention} that takes in a context $c_t=(e_{t-L+1}, \ldots, e_t)$ of the $L$ most recent embedded observations as input. At timestep $t$, we embed the observation $o_t$ into $e_t$ with a linear layer. The context $c_t$ is stored in a sliding buffer and passed through multiple self-attention and feed-forward layers with residual connections. Since each token in the context attends to every other token, the contextualized sequence of output vectors encapsulates information about the recent history of $e_t$. Appendix G offers a gradient-dynamics interpretation of our transformer-based in-context learning and its relations with other in-context RL architectures.

We take the final output vector $z_t$, corresponding to the most recent observation token within $c_t$, as the IA's state representation. This allows the IA to use its observational history when choosing what information to convey to the CA. The emitted message from the IA is generated by another linear layer with a $\tanh$ activation: $m_t = \tanh(f_{\phi}^{\text{Msg}}(z_t))$. The $\tanh$ non-linearity is used to constrain the resulting message vector. The IA is not trained on any direct return-maximization objective; no policy gradient flows from the CA into the IA.

The Control Agent (CA) is parameterized by $\theta$ and is tasked with controlling actions. It operates as a standard RL agent whose policy $\pi_{\theta}(\cdot|o_t, m_t)$ and value function $V_{\theta}(o_t, m_t)$ are conditioned on both the observation and the message. While CORAL is algorithm-agnostic, we instantiate the CA with Proximal Policy Optimization (PPO)~\cite{schulman2017proximal} for our primary online adaptation experiments. Its objective is to maximize the task reward.

\subsection{Pre-train the Communicative Representation}

The agents' parameters $\phi,\theta$ are optimized concurrently during pre-training. The learning objectives are constructed so as to encourage an IA that captures the world's dynamics and can communicate that knowledge, and a CA that can use that communication to solve the task.

\subsubsection{The Information Agent as a Communicative World Model}

The IA is trained via a composite self-supervised loss function, $\mathcal{L}(\phi)$, that does not depend directly on the extrinsic task reward. It combines objectives for world dynamics modeling, message consistency, and communicative efficacy.
\begin{equation*}
    \mathcal{L}(\phi) = \lambda_{\text{Dyn}} \mathcal{L}_{\text{Dyn}}(\phi) + \lambda_{\text{Coh}} \mathcal{L}_{\text{Coh}}(\phi) + \lambda_{\text{Causal}} \mathcal{L}_{\text{Causal}}(\phi).
\end{equation*}

\textbf{Dynamics Awareness Loss ($\mathcal{L}_{\text{Dyn}}$):} This objective anchors the IA's representations by learning to anticipate the results of the CA's actions. From the message $m_t$ and action $a_t$ selected by the CA, the IA predicts the next observation $\hat{o}_{t+1}$, reward $\hat{r}_{t+1}$, and termination probability $\hat{d}_{t+1}$. The loss $\mathcal{L}_{\text{Dyn}}$ is given below, and see Appendix C for more details. 
\begin{align*}
\scriptsize
\mathbb{E}_t [  {\| \hat{o}_{t+1} - o_{t+1} \|^2} + {(\hat{r}_{t+1} - r_{t+1})^2}   +  \text{BCE}(\hat{d}_{t+1}, d_{t+1})], 
\end{align*}
where predictions are generated by prediction heads ($f_{\phi}^{\text{Dyn}}$) conditioned on the message $m_t$ and action $a_t$: $(\hat{o}'_{t+1}, \hat{r}_{t+1}, \hat{d}_{t+1}) = f_{\phi}^{\text{Dyn}}(m_t, a_t)$.The first two terms in the loss use mean-squared error since the observation and reward predictions are continuous, whereas the last term uses binary cross-entropy (BCE) for probabilistic outputs.

\textbf{Temporal Coherence Loss ($\mathcal{L}_{\text{Coh}}$):} To promote temporal coherence in communication, the IA is also trained to 
predict the actual next-step message
$m_{t+1}$ based on the current message $m_t$ through a specific prediction head ($f_{\phi}^{\text{MsgPred}}$) with the loss given by $\mathcal{L}_{\text{Coh}} = \mathbb{E}_t \left[ \| \hat{m}_{t+1} - m_{t+1} \|^2 \right]$, where $\hat{m}_{t+1} = f_{\phi}^{\text{MsgPred}}(m_t, a_t)$. We encourage the message to be a compact representation of the state from which future states (and thus future messages) can be inferred. 

\begin{figure*}[ht]
    \centering
    \includegraphics[width=0.885\linewidth]{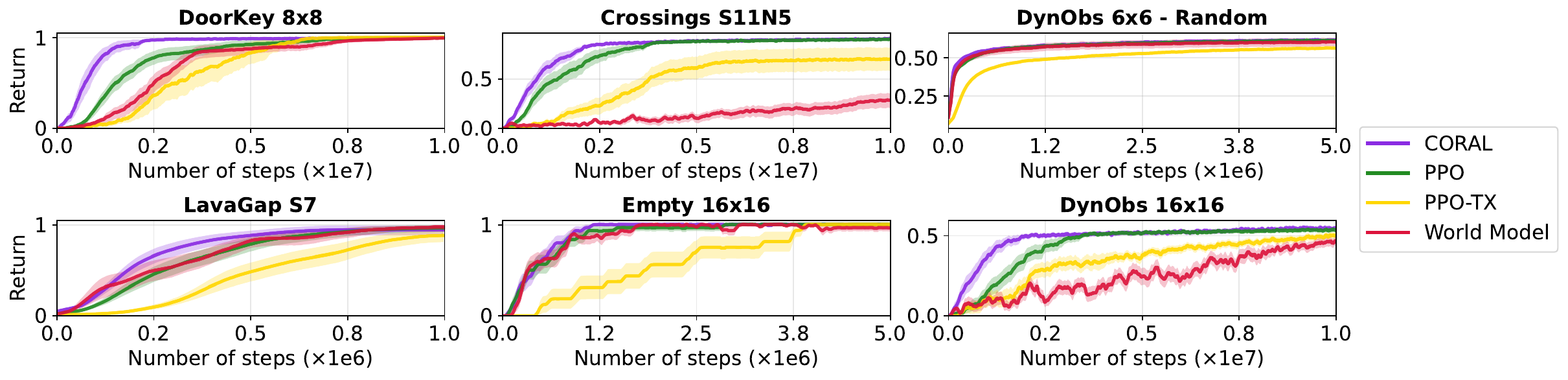}
    \caption{In-context adaptation with a pre-trained Information Agent. Learning curves show the mean episodic return ($\pm95\%$ CI) across 30  seeds for a randomly initialized Control Agent paired with a pre-trained, frozen CORAL IA. \textcolor[HTML]{923BE5}{CORAL} demonstrates higher sample efficiency and asymptotic performance compared to a \textcolor[HTML]{339332}{standard PPO}, \textcolor[HTML]{FFD700}{PPO-Transformer}, and an \textcolor[HTML]{DD264A}{equivalent World Model} across unseen MiniGrid environments.}
    \label{fig:learning-curves}
\end{figure*}

\textbf{Causal Influence Loss ($\mathcal{L}_{\text{Causal}}$):} To shape the communication to be effective from the CA's perspective, we introduce an information-theoretic objective inspired by the work on measuring causal influence in emergent communication~\cite{lowe2019pitfalls}. This loss encourages the IA to produce messages that cause a beneficial and decisive shift in the CA's policy. We quantify this per-step shift using a metric we term the Instantaneous Causal Effect (ICE), defined as the Reverse KL Divergence between the CA's policy with and without the message:
\begin{align}
    \text{ICE}_t =D_{\text{KL}}\Big(\pi_{\theta}(\cdot|o_t, \mathbf{0}) \parallel \pi_{\theta}(\cdot|o_t, m_t)\Big). \tag{ICE}
    \label{eq:ice}
\end{align}
Here, $\pi_{\theta}(\cdot|o_t, \mathbf{0})$ is conditioned on a zero-vector message, representing a non-informative communicative input.

To encourage helpful messages from IA that yield high-return actions, we multiply the ICE by a hybrid utility signal $\mathcal{U}_t$. The signal combines two sources of utility: The first is the Advantage estimate $A_t$ (e.g., GAE~\cite{schulman2015high}), which provides a low-variance estimate of the long-term benefit of taking an action. It is calculated as the exponentially-weighted average of temporal difference errors: $A_t = \sum_{k=0}^{\infty} (\gamma\lambda)^k \delta_{t+k}$, where $\delta_t = r_{t+1} + \gamma V_{t+1} - V_t$. The second source is the immediate change in the CA's own value estimate, $\Delta V_t = V_{\theta}(o_t, m_t) - V_{\theta}(o_t, \mathbf{0})$. To ensure that both signals contribute on a comparable scale regardless of reward magnitude or the stage of training, we apply z-score normalization, a standard technique for stabilizing policy gradient updates~\cite{schulman2017proximal,ioffe2015batch}. These normalized signals are then combined and clipped to form the final positive utility signal $\mathcal{U}_t = \max\{0, \alpha \cdot \text{norm}(\Delta V_t) + (1-\alpha) \cdot \text{norm}(A_t)\}$.
The final loss term maximizes the utility-weighted KL divergence over trajectories $\tau$ from the joint policy,
$\mathcal{L}_{\text{Causal}}(\phi) = - \mathbb{E}_{\tau \sim \pi_{\theta, \phi}} \left[\mbox{ } \mathcal{U}_t \cdot \text{ICE}_t\mbox{ } \right]$.

The utility signal $\mathcal{U}_t$ is treated as a constant during the optimization of $\phi$ by detaching it from the computation graph. This guarantees that the gradient only flows through the KL divergence term, correctly isolating the effect of rewarding the IA for how its message $m_t$ influences the CA's policy, rather than for influencing the value estimates that comprise the utility signal itself.

\subsection{Multi-Environment Training for Generalization}
CORAL aims to produce a communicative prior that is not overfitted to a single task but is instead broadly applicable. It is well-established that deep RL agents can achieve high performance by overfitting to the specific stochasticity of their training environments, yet fail to generalize to slightly different, unseen situations \cite{cobbe2020leveraging}.

To directly address this challenge, our pre-training regime moves beyond single-task optimization and instead trains the CORAL agents on a diverse, discrete set of tasks, $\mathcal{T} = \{\text{env}_1, \ldots, \text{env}_K\}$, that share common entities and dynamics but require different solutions. Our implementation is based on modern, vectorized training within a single process. Inspired by the architectural principles of large-scale distributed agents \cite{espeholt2018impala} and using an implementation pattern similar to high-performance JAX-native frameworks like PureJaxRL \cite{lu2022discovered}, we leverage the \texttt{vmap} transformation in JAX \cite{jax2018github} to run N parallel environments simultaneously on a single accelerator.

At the beginning of each rollout, each of the $N$ parallel environment instances is randomly assigned a task from our distribution $\mathcal{T}$. This ensures that every gradient update batch contains a rich mixture of experiences from across the task family. By training on this diverse data stream, we prevent catastrophic forgetting and force the Information Agent to learn an abstract dynamics model, capturing the fundamental rules and entities common across environments rather than memorizing the specifics of any single instance. This multi-task regime is critical for learning a communicative representation that can serve as a truly generalizable prior.

\subsection{Deployment for Rapid In-Context Adaptation}
At deployment time, we fix the parameters $\phi$ of the Information Agent. IA effectively becomes a frozen deterministic module providing a continuous stream of contextual information. As such, we can precisely evaluate our main hypothesis that a trained communication protocol allows the Control Agent to adapt fast and with few samples.

We formalize this evaluation through the lens of in-context reinforcement learning \cite{lee2023supervised,laskin2023incontext}. The Control Agent learning can be cast as inference of optimal policy in a new environment conditioned on the history of observations and messages constituting the context.

\section{Experimental Results}

\begin{table*}[ht]
\scriptsize
\centering
  \begin{tabular}{l c cc cc cc cc}
    \toprule
    \multirow{2}{*}{Environment} &
    \multirow{2}{*}{Max. Perf} &
    \multicolumn{2}{c}{\textcolor[HTML]{923BE5}{CORAL}} &
    \multicolumn{2}{c}{\textcolor[HTML]{339332}{PPO}} &
    \multicolumn{2}{c}{\textcolor[HTML]{FFD700}{PPO-TX}} &
    \multicolumn{2}{c}{\textcolor[HTML]{DD264A}{World Model}} \\
      & & TTT & SR & TTT & SR & TTT & SR & TTT & SR \\
    \midrule
DoorKey 8x8 & 1.00 & \textbf{1.0 $\pm$ 0.2$^{\dag}$} & 100\% & 2.3 $\pm$ 0.3 & 100\% & 3.9 $\pm$ 0.7 & 100\% & 3.0 $\pm$ 0.4 & 89\% \\
Crossings S11N5 & 1.00 & \textbf{1.3 $\pm$ 0.1$^{\dag}$} & 100\% & 1.9 $\pm$ 0.3 & 100\% & 3.9 $\pm$ 0.4 & 73\% & 6.9 $\pm$ 1.2 & 19\% \\
DynObs 6x6 - Random & 0.74 & 1.5 $\pm$ 0.2 & 100\% & 1.5 $\pm$ 0.1 & 97\% & 4.4 $\pm$ 0.2 & 36\% & \textbf{1.3 $\pm$ 0.1} & 97\% \\
LavaGap S7 & 1.00 & \textbf{0.3 $\pm$ 0.0$^{\dag}$} & 95\% & 0.4 $\pm$ 0.0 & 95\% & 0.6 $\pm$ 0.1 & 79\% & 0.4 $\pm$ 0.0 & 94\% \\
Empty 16x16 & 1.00 & \textbf{0.5 $\pm$ 0.1} & 100\% & 0.7 $\pm$ 0.1 & 100\% & 2.0 $\pm$ 0.3 & 100\% & \textbf{0.5 $\pm$ 0.1} & 100\% \\
DynObs 16x16 & 0.85 & \textbf{7.0 $\pm$ 0.6$^{\dag}$} & 45\% & 7.9 $\pm$ 0.5 & 36\% &  8.2 $\pm$ 0.2 & 30\% & 8.5 $\pm$ 0.7 & 30\% \\
    \bottomrule
  \end{tabular}
\caption{
Time-to-threshold analysis (TTT, millions of steps) $\pm$ 95\% confidence interval (CI) to reach 90\% of maximum performance; SR: success rate. \textbf{Bold}: best per environment. ${\dag}$: significant improvement over next-best (Welch's t-test, $p < 0.05$).
}
\label{table: TTT}
\end{table*}

We validate CORAL across three distinct regimes to assess its generalization capabilities across different data modalities. First, using the \texttt{Navix} grid-world suite~\cite{pignatelli2024navix}, we assess \textbf{Online Adaptation}, measuring the agent's ability to learn unseen tasks from scratch, and \textbf{Zero-Shot Transfer}, evaluating robustness to increased task complexity without further training. Second, we benchmark \textbf{Offline Generalization} on the D4RL continuous control suite~\cite{fu2020d4rl}. Our analysis compares CORAL against standard PPO and monolithic World Model baselines in the online settings, while extending the comparison to state-of-the-art In-Context RL methods, including Decision Transformer (DT) and Agentic Transformer (AT), for the offline benchmarks.

\subsection{Accelerated Control Learning via In-Context Communication}

\paragraph{Protocol.} To benchmark online in-context adaptation, we isolate source and target environments during adaptation. In this Accelerated Adaptation regime, we train an Information Agent as a single foundation model on a distribution of source tasks (\textit{e.g.}, \textit{navigation, lava, object-centric, dynamic-obstacle, etc.}), then freeze its weights (see Appendix C.1 for the complete pre-training set of environments). Next, we place a fresh, randomly initialized Control Agent into a distribution of unseen target tasks (\textit{e.g.}, \textit{DoorKey-8x8}) which learns to solve the task by consuming the frozen IA's communicative context alone. This clearly isolates the sample-efficiency boost granted by the communicative prior, as the agent must learn from zero examples. This contrasts with our Zero-Shot Generalization section, which examines fixed policies with no further adaptation.

Our first set of experiments analyzes whether a frozen pre-trained CORAL IA can accelerate the learning of a randomly initialized CA from scratch on new tasks. We randomly initialize a new CA from scratch and evaluate it together with the frozen IA in previously unseen environments. 

Fig.~\ref{fig:learning-curves} demonstrates a large improvement in sample efficiency. The CORAL agent is able to achieve near-optimal rewards roughly twice as fast as the PPO agent in environments like \textit{DoorKey 8x8} and \textit{Crossings S11N5}. This improvement is even more dramatic when we begin to evaluate our method on harder or larger variants of the training environments. Consider Dynamic Obstacles 16x16 (\textit{DynObs 16x16}), a much larger and more complicated environment than the environment of the same nature seen by agents during pretraining. While the baseline improves marginally over Random performance, CORAL is able to bootstrap the CA to effectively solve the task. This suggests that the learned communicative prior can provide strong yet robust guidance for the agent to avoid the exploration difficulty posed by these sparse-reward navigation tasks, even as the problem scales up.

To better quantify this improvement in sample efficiency, we conduct a time-to-threshold (TTT) analysis on each environment. We calculate the mean number of environment timesteps it takes each method to reach 90\% of that agent's maximum achievable performance in each environment. We report TTT results in Table~\ref{table: TTT}. As shown, CORAL is able to meet this performance threshold significantly faster than either baseline across most environments. CORAL reaches our target performance after significantly fewer timesteps in \textit{Doorkey 8x8}, 2.3 and 3 times faster than PPO and the World Model baseline, respectively. Similarly, we see a 1.5-fold and 5+ fold improvement in TTT over PPO and the World Model baseline in \textit{Crossings S11N5}. Lastly, we report the success rate (SR) to show CORAL's increased likelihood of actually solving these tasks. CORAL shows a 25\% relative improvement in SR over PPO and 50\% over WM in \textit{DynObs16x16}, the single environment where WM performed the best relative to PPO.

These consistent results indicate that the improvement from our framework stems from the fact that it allows agents to communicate rather than using a different architecture. The gap in performance between CORAL and PPO highlights the strong learning signal provided by the message stream for agent exploration in sparse-reward environments. Moreover, CORAL significantly outperforms the architecturally equivalent World Model baseline, also highlighting the importance of our pre-training strategy. By pre-training the Information Agent with self-supervised objectives that do not directly depend on maximizing task reward, we are able to align the CA and IA onto a communication protocol that is more general than the task-specific representations learned by WM.

\subsection{Integrated In-Context Communication and Control for Zero-Shot Generalization}

Beyond accelerating learning from scratch, we investigate the robustness of the learned policies in a challenging zero-shot transfer setting where no further training is allowed. For this evaluation, we use the same generalist Information Agent pre-trained on our full task distribution $\mathcal{T}$. We then pre-train the CORAL CA and baselines on a specific, simpler source task (e.g., \textit{DoorKey6x6}). For the Decision Transformer (DT) baseline, we trained the model on offline trajectories collected from the converged PPO policy on these source tasks. Finally, we freeze all agent parameters and evaluate their performance on more complex, unseen target environments (e.g., \textit{DoorKey8x8}). These experiments measure how well the learned policy generalizes to configurations of increased scale and complexity.

As shown in Table~\ref{table: MaxPerGeneralization}, we see that the CORAL architecture allows for strong zero-shot generalization performance in comparison to our baselines. The CORAL agent that was pre-trained on both \textit{DoorKey6x6} and \textit{LavaGapS6} obtains a much higher average return than both the PPO and World Model agents when placed in the more challenging environments of \textit{DoorKey8x8} and \textit{LavaGapS7}, respectively. CORAL also outperforms the DT, suggesting that the communicative prior generalizes better to dynamic shifts than purely sequence-based cloning of source task demonstrations in ICRL settings.

\begin{table*}[ht]
\centering
\scriptsize
\begin{tabular}{@{}lccccc@{}}
\toprule
\textbf{Environment} & \textcolor[HTML]{923BE5}{\textbf{CORAL}} & \textcolor[HTML]{339332}{PPO} & \textcolor[HTML]{FFD700}{PPO-TX} &\textcolor[HTML]{DD264A}{WM} & DT \\
\midrule 
DoorKey8x8 &  $\mathbf{0.95\pm9e^{-4}}^{\dag}$ & $0.78\pm6e^{-4}$ & $0.81\pm7e^{-4}$& $0.86\pm7e^{-3}$ & $0.84\pm1e^{-3}$\\
CrossingsS11N5 & $0.45\pm9e^{-3}$ & $0.48\pm8e^{-3}$ & $\mathbf{0.49\pm6e^{-3}}$ & $0.20\pm5e^{-3}$ &$0.43\pm8e^{-3}$\\
DynObs6x6-Random & $\mathbf{0.64\pm2e^{-3}}^{\dag}$ & $0.43\pm2e^{-3}$ & $0.54\pm4e^{-3}$ & $0.33\pm7e^{-3}$ & $0.61 \pm 3e^{-3}$ \\
LavaGapS7 & $\mathbf{0.77\pm6e^{-3}}^{\dag}$ & $0.63 \pm 6e^{-3}$ & $0.63 \pm 5e^{-3}$ & $0.65\pm5e^{-3}$& $0.74 \pm 9e^{-3}$\\
Empty16x16 & $\mathbf{0.90\pm 4e^{-3}}$ & $0.86\pm 3e^{-3}$ & $0.87 \pm 5e^{-3}$ & $0.88\pm4e^{-3}$ & $\mathbf{0.90\pm 5e^{-3}}$ \\
DynObs16x16 & $\mathbf{0.52\pm6e^{-3}}^{\dag}$ & $0.50\pm5e^{-3}$ & $0.47\pm8e^{-3}$ & $0.16\pm6e^{-3}$ & $0.43 \pm4e^{-3}$ \\
\bottomrule
\end{tabular}
\caption{Zero-shot performance with frozen weights (pre-trained on simpler source tasks). Mean episodic return $\pm$ 95\% CI over 1M steps.}
\label{table: MaxPerGeneralization}
\end{table*}

\begin{table*}[ht]
\scriptsize
\centering
\begin{tabular}{ll|cccc|c}
\hline\hline
Dataset & Environment & TD3+BC & DT & AT & AD & \textcolor[HTML]{923BE5} {\textbf{CORAL}} \\ \hline
Medium-Expert & HalfCheetah & 96.59 & 93.40 & $95.81 \pm 0.25$ & $94.21 \pm 0.46$ & \textbf{96.67} $\pm$ \textbf{0.24} \\
Medium-Expert & Hopper      & 113.22 & 111.18 & $115.92 \pm 1.26$ & $108.32 \pm 0.95$ & \textbf{116.06} $\pm$ \textbf{1.02}\\
Medium-Expert & Walker      & 112.21 & 108.71 & \textbf{114.87} $\pm$ \textbf{0.56} & $111.36 \pm 0.46$ & $114.48 \pm 0.65$ \\ \hline
Medium        & HalfCheetah & 48.93 & 42.73 & $45.12 \pm 0.34$ & $42.28 \pm 1.18$ & \textbf{49.01} $\pm$ \textbf{0.29} \\
Medium        & Hopper      & 70.44 & 69.42 & $70.45 \pm 0.45$ & 72.58 $\pm$ 0.54 & \textbf{73.15} $\pm$ \textbf{0.48} \\
Medium        & Walker      & 86.91 & 74.70 & $88.71 \pm 0.55$ & $85.96 \pm 0.46$ & \textbf{88.86} $\pm$ \textbf{0.41} \\ \hline
Medium-Replay & HalfCheetah & 45.84 & 40.31 & \textbf{46.86} $\pm$ \textbf{0.33} & $41.28 \pm 0.21$ & $45.91 \pm 0.37$ \\
Medium-Replay & Hopper      & 98.12 & 88.74 & $96.85 \pm 0.41$ & $91.32 \pm 0.66$ & \textbf{98.33} $\pm$ \textbf{0.72}\\
Medium-Replay & Walker      & 91.17 & 68.22 & \textbf{92.32} $\pm$ \textbf{1.21} & $89.21 \pm 1.42$ & 86.74 $\pm$ 1.13\\ \hline
\multicolumn{2}{c|}{Total Average} & $84.83$ & $77.49$ & $85.21$ & $82.84 $ & \textbf{85.47}\\ \hline\hline
\end{tabular}
\caption{Normalized scores on D4RL continuous control tasks. CORAL is trained fully offline.}
\label{d4rl}
\end{table*}

\subsection{D4RL Results}
Beyond online interaction, we benchmark CORAL against D4RL~\cite{fu2020d4rl}. As per the standard offline ICRL setup~\cite{moeini2025survey,chen2021dt}, we approach it as completely offline. In this setting, the IA learns about world dynamics and reward from the offline transitions alone, creating a distilled dataset as a communicative prior. Meanwhile, the CA learns a message-conditioned policy concurrently. Table~\ref{d4rl} displays CORAL's total average normalized score of \textbf{85.47}, beating the DT (77.49) and achieving comparable performance to TD3+BC (84.83), and the Agentic Transformer (AT)~\cite{liu2023emergent}. CORAL also notably outperforms other algorithms on the difficult \textit{Medium-Replay} datasets (\textbf{98.33} on Hopper-M-Replay vs. 88.74 for DT and AT's 96.85), indicating CORAL's communicative prior is able to discard noise from sub-optimal demonstrations.

\subsection{Emergent Communicative Protocol Analysis}

To verify that communication is causally responsible for the observed performance gains, we study the protocol's Instantaneous Causal Effect (Eq.~\ref{eq:ice}). A high ICE value indicates that the message is causing a decisive shift in the CA's behavior.

\begin{figure}[ht]
    \centering    \includegraphics[width=1\linewidth]{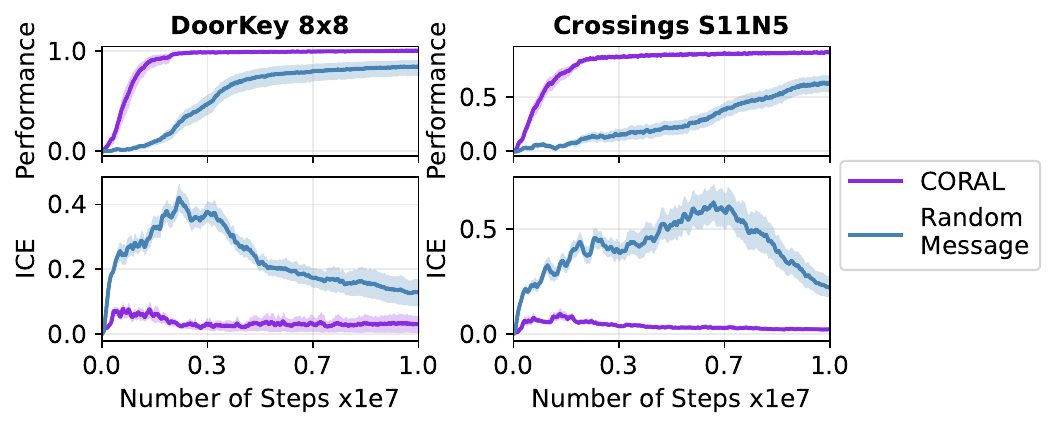}
    \caption{Communication Impact on Adaptation. Performance (top) and corresponding ICE metric (bottom) for unseen tasks.}
    \label{fig:ice-compact}
\end{figure}

Fig.~\ref{fig:ice-compact} shows the mean and 95\% CI of ICE averaged across all steps of CORAL and one CA trained with random messages. Although these random messages provide maximal ICE by completely distracting the agent, they provide no beneficial learning signal. Instead, CORAL's ICE increases with performance as the CA learns to trust the IA, and decreases as the learned policy is fixed and the message becomes expected information. The shape of ICE going high when learning and low when converged suggests that the CORAL protocol provides learning regularization by only asserting a strong influence when the agent is uncertain.

\subsection{Ablation Study}
We investigate the contribution of our specific training objectives to the emergent communication protocol (see Appendix F for full training curves).

\begin{figure}[!t]
    \centering
    \includegraphics[width=0.9815\linewidth]{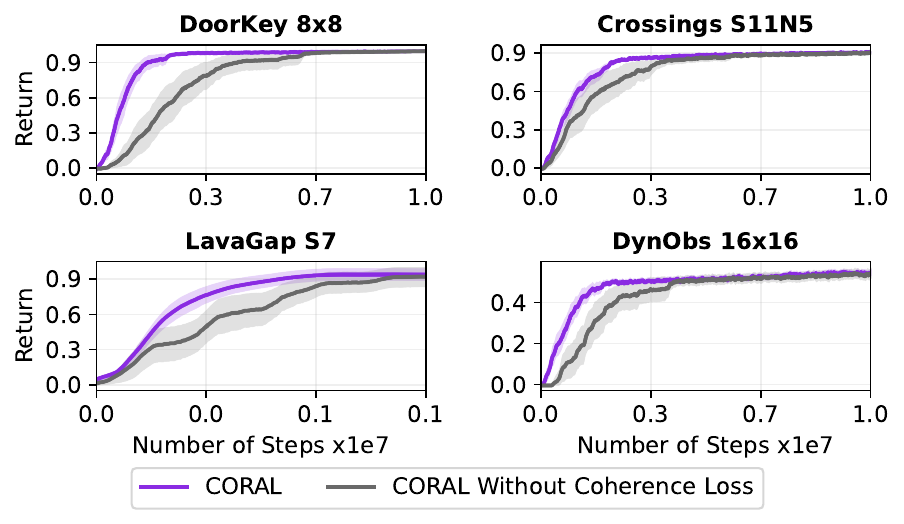}
    \caption{Ablation of $\mathcal{L}_{\text{Coh}}$. Comparison between \textcolor[HTML]{8A2BE2}{CORAL} and an ablation variant \textcolor[HTML]{696969}{without coherence loss}.
    }
    \label{fig:ablation-coherence}
\end{figure}

\paragraph{Effect of Coherence Loss.} We hypothesized that smooth transitions in the message space reduce the variance in the CA's value estimation. Our tests, in Fig.~\ref{fig:ablation-coherence}, confirm that $\mathcal{L}_{\text{Coh}}$ is an essential regularizer. Without it leads to high variance, slower learning, and poorer performance across all environments.

\begin{figure}[!t]
    \centering
    \includegraphics[width=0.9815\linewidth]{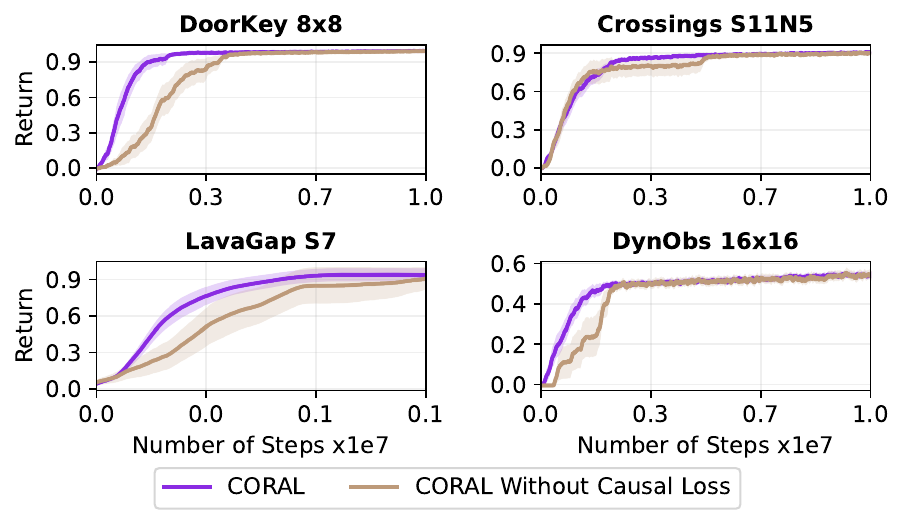}
        \caption{Ablation of $\mathcal{L}_{\text{Causal}}$. Comparison between \textcolor[HTML]{8A2BE2}{CORAL} and an ablation variant \textcolor[HTML]{BD9A7A}{without causal loss}.
        }
    \label{fig:ablation-causal}
\end{figure}

\paragraph{Effect of Causal Influence Loss.} We also ablate the requirement for the Causal Influence objective by training a version of CORAL without incentivizing the IA to influence the CA's policy. As shown in Fig.~\ref{fig:ablation-causal}, removing $\mathcal{L}_{\text{Causal}}$ hurts sample efficiency and final performance in some cases. This shows that having only a ``passive" world model trained to describe the dynamics faithfully is not enough to boost downstream tasks. The causal loss incentivizes pragmatic communication where only messages that cause beneficial shifts in policy are communicated, rather than wastefully encoding easily memorizable but ineffective static environmental features.

\section{Related Works}
\label{sec:related}
\paragraph{In-Context RL.} In-context learning, a concept recently popularized by large language models (LLMs) research and applications \cite{dong-etal-2024-survey}, refers to the ability to learn from a few examples in the context and adapt without changing model weights. The idea of few-shot adaptation in ICRL aligns with earlier work on gradient-free meta learning, where recurrent neural networks, aiming to extract task-specific hidden context from historical interactions, are incorporated into RL training \cite{botvinick16learnRL,abbeel16rl2}. Recent advancements on ICRL leverage transformers and LLMs to extract context variables (e.g., hidden states) from historical trajectories~\cite{lee2023supervised,krishnamurthy2024can}. Pre-training is key to the success in in-context adaptation. Most existing works fall within the categories of (self-)supervised pre-training and reinforcement pre-training~\cite{moeini2025survey}.  

Supervised pre-training, similar to imitation learning, encourages agents to find trajectories in the offline data that are similar to the current test task and to imitate offline actions for better generalization \cite{raparthy24icrl,xu22prompt-dt}. Self-supervised approaches, such as decision transformers \cite{chen2021dt,liu2023emergent,huang2024decision}, leverage the transformer's autoregressive generation to predict the next action based on state observations and reward feedback. Even though some advanced techniques, such as hindsight information matching \cite{furuta21generalDT,tao23sce}, help improve out-of-distribution generalization, the (self-)supervised approaches largely depend on the quality of the offline data. 

Our proposed CORAL belongs to the class of reinforcement pre-training, where the policy is not purely extracted from static, offline data but also from live interactions with a variety of environments. Compared with existing works on transformer-based ICRL \cite{bauer23,grigsby2024amago,wang2024transformers}, which integrate context learning into policy learning, our work further decouples the two learning processes and adopts a \textit{hybrid} approach that combines supervised pre-training for world model training and reinforcement pre-training for control policies. 

\paragraph{World Models.} World models (WMs), typically represented by generative neural networks, serve as the agents' mental models of the world based on what they can perceive, which helps agents to conceive future consequences of their actions, thus planning in sequential decision-making~\cite{schmidhuber18wm}. Some early efforts focus on building visual WMs using autoencoders and recurrent neural networks~\cite{schmidhuber18wm,hafner19a,hafnerdream,hafner25worldmodel}, where WMs predict the next latent representation for planning purposes based on those of previous image inputs. In more structured tasks (e.g., Go),  simple predictive models plus Monte Carlo tree search (MCTS) also yield superior performance~\cite{silver20wm,silver21on-off-wm,kim-tao25col}. 

Our CORAL aligns with the recent trend of transformer-based WMs~\cite{micheli2023transformers,robine2023transformerbased} for its improved sample efficiency, compared with RNN-based ones. However, these transformer WMs, like most WMs, directly incorporate the general-purpose dynamics modeling learning with task-specific RL, where agents treat the transformer-generated latent representations as the new state variables.  In contrast, our approach decouples the two learning tasks, and the transformer is only tasked with understanding the shared environment dynamics and communicating them to the control agent as side information or context. Most relevant to ours is \cite{toledocodreamer}, which studies communication-based decentralized WMs. This work proposes to equip each agent with a WM in a multi-agent RL task, where WMs first communicate with each other on future predictions. CORAL, instead, focuses on tackling a single-agent RL from an emergent communication perspective, where the transformer WM is still the mental model without actually optimizing the policy.

\section{Conclusion}

This work introduces CORAL, a framework for learning a communicative prior for in-context reinforcement learning. By pre-training an Information Agent as a communicative world model with objectives decoupled from direct task rewards, we learn a reusable protocol that effectively conveys relevant knowledge to instruct a control agent's actions. Our experiments show that using this pre-trained communicative prior enables improvements in sample efficiency for several online adaptation and better zero-shot generalization in unseen, sparse-reward tasks. Additionally, we show CORAL achieves competitive results on offline RL tasks, suggesting emergent communication is promising not just for coordination but as a general method for rapid adaptation.


\ifwithappendix
\fi

\bibliographystyle{named}
\bibliography{ijcai26}

\ifwithappendix

\clearpage
\begin{appendix}
    
\setcounter{secnumdepth}{2}
\section*{Appendix}

\section{Code availability}
\label{appendix:code_availability}
The complete source code for CORAL and the experiments presented in this paper is available at \url{https://github.com/fernando-ml/CORAL} to ensure reproducibility. Our implementation is built in JAX \cite{jax2018github} and leverages significant components, particularly for environment vectorization and training loops, from the PureJaxRL framework \cite{lu2022discovered}\footnote{\url{https://github.com/luchris429/purejaxrl/tree/main}} and Navix \cite{pignatelli2024navix}\footnote{\url{https://github.com/epignatelli/navix}}. While core architectural choices (Appendix~\ref{appendix:hyperparameters_and_setup}) and most hyperparameters for baseline algorithms are consistent with PureJaxRL defaults for Navix. D4RL baselines and CORAL's code follow JAX-CORL repository\footnote{\url{https://github.com/nissymori/JAX-CORL/tree/main}}.

\section{Limitations and Future Work}
While our work demonstrates the significant potential of using pre-trained communicative priors for in-context adaptation, it also highlights several avenues for future extension. 

First, an interesting direction could be related to the \textbf{scalability and design of the communication protocol itself}. CORAL uses a fixed-dimensional dense vector as its protocol. While we find that this is expressive enough to work well in the domains we study, better expressive or more parameter-efficient protocols may be possible, especially in settings with higher-dimensional inputs, requiring more compositional reasoning. Structured protocols with discrete tokens drawn from some learned vocabulary could yield improvements in expressivity. Further, our framework only considers the ``cheap talk'' setting where communication between agents is free ~\cite{crawford82signaling,lowe2019pitfalls,tao23pot}. Adding a cost to sending messages is a natural extension, which could incentivize the IA to learn even more efficient sparse protocols, only speaking when its expected utility from doing so exceeds its expected cost.

A second direction for future work is centered around \textbf{improving the pre-training distribution with Unsupervised Environment Design (UED)}. The generalization potential of the CORAL Information Agent is inherently bounded by the structural diversity of its pre-training distribution. One exciting direction is to leverage automated curriculum generation techniques to iteratively produce harder environments for the agent to train from. Automatically generating levels that are difficult for the IA to transfer to, i.e., maximizing regret, would allow us to produce a more robust communicative prior that generalizes to important edge cases.

Finally, our framework can be extended to more \textbf{complex multi-agent topologies}. To better understand the impact of pre-training with a communicative prior, this work focused on a simple asymmetric pair of agents. A natural extension is generalizing this framework to cases where a single Information Agent learns to communicate with multiple, possibly specialized, Control Agents. We could also consider settings where agents can communicate bi-directionally, and all agents are homogeneous speakers/listeners.

\section{Experiment Setup and Hyperparameters}
\label{appendix:hyperparameters_and_setup}

\paragraph{Software.} We used the following software versions:
\begin{itemize}
    \item Python 3.10.18 - Python Software License \url{https://docs.python.org/3/license.html}
    \item CUDA 12.4 - NVIDIA Software License Agreement \url{https://docs.nvidia.com/cuda/eula/index.html}
    \item Jax 0.5.3 - Apache License 2.0 \url{https://github.com/jax-ml/jax}
    \item Flashbax 0.1.2 - Apache License 2.0 \url{https://github.com/instadeepai/flashbax}
    \item Chex 0.1.89 - Apache License 2.0 \url{https://github.com/google-deepmind/chex}
    \item Optax 0.2.4 - Apache License 2.0 \url{https://github.com/google-deepmind/optax}
    \item Flax 0.10.4 - Apache License 2.0 \url{https://github.com/google/flax}
    \item Navix 0.7.0 - Apache License 2.0 \url{https://github.com/epignatelli/navix}
    \item Gymnax 0.0.8 - Apache License 2.0 \url{https://github.com/RobertTLange/gymnax}
    \item PureJaxRL - Apache License 2.0 \url{https://github.com/luchris429/purejaxrl}
\end{itemize}

\paragraph{Hardware.} All experiments were conducted on NVIDIA Tesla V100-PCIE-32GB GPUs. A typical experimental run, consisting of training one algorithm for $10^7$ total environment time steps on a Navix environment, completed in approximately 12 to 26 minutes over 30 seeds.

\subsection{Pre-training Environment Distribution}
\label{appendix:pretraining-set}

To foster the development of a generalizable transferable communication protocol, our pre-training stage utilizes a diverse distribution of tasks, $\mathcal{T}$. The environments were selected to expose the Information Agent (IA) to a wide range of mechanics, entities, and objectives. This multi-world regime is designed to drive the IA to learn a model of grid-world dynamics, rather than overfitting to the specifics of a single task. 
The pre-training distribution consists of the following environments:

\begin{itemize}
    \item \textbf{Empty-Random-8x8}: The simplest task, it requires the agent to navigate to a goal in an empty room with randomized start and goal positions. It serves to teach the IA concepts of navigation and goal-directedness in the absence of other factors.
    \item \textbf{GoToDoor8x8}: This task requires the agent to navigate to a specific object (the door) rather than a generic goal tile. This teaches the IA the concept of object-centric goals so it can communicate information about specific entities in the environment.
    \item \textbf{FourRooms}: In this environment, the agent needs to navigate through multiple interconnected rooms. This tests the IA's ability to model and communicate information relevant to planning and navigation with multi-room layouts.
    \item \textbf{CrossingS9N3}: In this environment, the agent must cross a room with static obstacles, teaching the IA to learning about path planning and the semantics of impassable objects like walls.
    \item \textbf{LavaGapS6}: It introduces a "dangerous" terrain type (lava) that the agent must avoid, introducing IA to the concept of environmental constraints and affordances.
    \item \textbf{Dynamic-Obstacles-5x5}: This environment features mobile obstacles (balls) that move randomly. Since this environment is non-stationary, the IA must develop a compelling understanding of the world and communicate about changing states.
    \item \textbf{DoorKey-Random-6x6}: This task introduces the concept of conditional objectives. Here, the agent must first find a key, pick it up, and then use it to open a locked door to reach the goal. For this environment, the IA needs to model a longer sequence of sub-tasks and dependencies between them, and communicate accordingly. 
\end{itemize}
The held-out target environments used for online adaptation and zero-shot evaluation are summarized in Table~\ref{tab:target_env_summary}.

\subsection{Neural Network Architectures and Hyperparameters}
CORAL's training and deployment setup and neural network designs are presented in Table \ref{tab:coral_hyperparams}, \ref{tab:coral_ia_architecture}, and \ref{tab:coral_ca_architecture}, respectively.
\begin{table}[!hb]
\centering
\caption{Train \& Deployment Hyperparameters for CORAL. Shared hyperparameters between IA and CA come from Navix's PureJaxRL example for PPO.}
\label{tab:coral_hyperparams}
\resizebox{\linewidth}{!}{
\begin{tabular}{@{}lc@{}}
\toprule
\textbf{Parameter} & \textbf{Value} \\
\midrule
\multicolumn{2}{@{}l}{\textit{ Configuration}} \\
\makecell[l]{Number of parallel environments \\ (Pre-train)} & 128 \\
\makecell[l]{Number of parallel environments \\ (Deploy)} & 16 \\
\makecell[l]{Total timesteps \\ (Pre-train)} & $5 \times 10^7$ \\
\makecell[l]{Total timesteps \\ (Deploy)} & $1 \times 10^7$ \\ 
Number of steps per rollout & 128 \\
Number of minibatches & 8 \\
Update epochs & 4 \\
\midrule
\multicolumn{2}{@{}l}{\textit{Learning Parameters}} \\
Optimizer & Adam \\
Learning rate (LR) & $2.5 \times 10^{-4}$ \\
Linear learning rate decay & True \\
Max gradient norm & 0.5 \\
Discount factor ($\gamma$) & 0.99 \\
GAE lambda ($\lambda$) & 0.95 \\
PPO clip epsilon & 0.2 \\
\midrule
\multicolumn{2}{@{}l}{\textit{Loss Coefficients}} \\
Entropy coefficient & 0.02 \\
Value function coefficient & 0.5 \\
Dyn. Awareness coefficient  $\lambda_{\text{Dyn}}$& 0.5 \\
Causal coefficient $\lambda_{\text{Causal}}$& 0.1 \\
Temporal Coherence coefficient $\lambda_{\text{Coh}}$ & 0.05 \\
Hybrid $\alpha$ for $\mathcal{U}_t$ & 0.5 \\
\midrule
\multicolumn{2}{@{}l}{\textit{Network Architecture}} \\
Hidden dimension & 128 \\
Message dimension & 32 \\
Activation function & tanh \\
\midrule
\multicolumn{2}{@{}l}{\textit{Transformer (Information Agent)}} \\
Context length & 4 \\
Number of attention heads & 4 \\
\midrule
\multicolumn{2}{@{}l}{\textit{Training Modes}} \\
Pretrain mode & Both IA and CA trained \\
Deploy mode & IA frozen, CA re-trained \\
\bottomrule
\end{tabular}
}
\end{table}

\begin{table*}[t]
\centering

\caption{Held-out Navix target environments and their relation to the IA pre-training distribution. Source environments are described in Appendix~\ref{appendix:pretraining-set}.}
\label{tab:target_env_summary}
\begin{tabular}{lcll}
\toprule
Target env. & Size & Related source coverage & Shift / mechanics \\
\midrule
DoorKey & \(8{\times}8\) & DoorKey-Random-6x6 & Larger key-door task \\
Crossing & S11N5 & CrossingS9N3 & Larger static-obstacle layout \\
DynObs-Random & \(6{\times}6\) & Dynamic-Obstacles-5x5 & Larger stochastic obstacles \\
LavaGap & S7 & LavaGapS6 & Larger lava-constraint task \\
Empty & \(16{\times}16\) & Empty-Random-8x8 & Larger navigation layout \\
DynObs & \(16{\times}16\) & Dynamic-Obstacles-5x5 & Much larger stochastic obstacles \\
\bottomrule
\end{tabular}
\end{table*}

\begin{table*}[!htbp]
\centering
\caption{Control Agent (CA) network architecture for CORAL. \texttt{Dense(in, out)} denotes a fully connected layer. The CA processes both partial observations and messages from the IA to produce policy and value estimates.}
\label{tab:coral_ca_architecture}
\begin{tabular}{@{}p{0.95\linewidth}@{}}
\toprule[\heavyrulewidth]
\textbf{CORAL Control Agent Architecture (Actor-Critic)} \\
\midrule
\\
\textbf{Variables:} \\
\texttt{obs\_dim} = Partial observation vector dimension (environment-dependent) \\
\texttt{action\_dim} = Number of discrete actions (7 for Navix environments) \\
\texttt{message\_dim} = Communication vector dimension \\
\texttt{hidden\_dim} = Latent representation dimension \\
\texttt{activation} = tanh \\
\\ \hline \\
\textbf{Control Agent (CA)} \\
\texttt{$\vartriangleright$\hspace{0.5em}Processes partial observation and message to produce policy and value estimates.} \\
\texttt{  obs\_layer = Dense(obs\_dim, hidden\_dim)} \\
\texttt{  msg\_layer = Dense(message\_dim, hidden\_dim)} \\
\texttt{  shared\_1 = Dense(hidden\_dim * 2, hidden\_dim)} \\
\texttt{  shared\_2 = Dense(hidden\_dim, hidden\_dim)} \\
\texttt{  actor\_head = Dense(hidden\_dim, action\_dim)} \\
\texttt{  critic\_head = Dense(hidden\_dim, 1)} \\
\\
\textbf{CA Forward Pass:} \\
\texttt{obs\_features = activation(obs\_layer(partial\_obs))} \\
\texttt{msg\_features = activation(msg\_layer(message))} \\
\texttt{combined = concat([obs\_features, msg\_features])} \\
\texttt{x = activation(shared\_1(combined))} \\
\texttt{x = activation(shared\_2(x))} \\
\texttt{policy = Categorical(actor\_head(x))} \\
\texttt{value = critic\_head(x)} \\
\texttt{return policy, value} \\
\\
\bottomrule[\heavyrulewidth]
\end{tabular}
\end{table*}

\begin{table*}[!htbp]
\centering
\caption{Information Agent (IA) network architecture for CORAL. \texttt{Dense(in, out)} denotes a fully connected layer. The IA processes partial observations through a transformer architecture to generate communication messages and world model predictions.}
\label{tab:coral_ia_architecture}
\begin{tabular}{@{}p{0.95\linewidth}@{}}
\toprule[\heavyrulewidth]
\textbf{CORAL Information Agent Architecture} \\
\midrule
\\
\textbf{Variables:} \\
\texttt{obs\_dim} = Partial observation vector dimension (environment-dependent) \\
\texttt{action\_dim} = Number of discrete actions (7 for Navix environments) \\
\texttt{message\_dim} = communication vector dimension \\
\texttt{hidden\_dim} = latent representation dimension \\
\texttt{context\_len} = transformer context buffer length \\
\texttt{num\_heads} = self-attention heads \\
\texttt{activation} = tanh \\
\\ \hline \\
\textbf{Information Agent (IA) - Message Generation} \\
\texttt{$\vartriangleright$\hspace{0.5em}Transformer-based agent processing partial observations to generate messages.} \\
\texttt{  obs\_tok = Dense(obs\_dim, hidden\_dim)} \\
\texttt{  pos\_embed = LearnedParameter(context\_len, hidden\_dim)} \\
\texttt{  ln1, ln2 = LayerNorm(), LayerNorm()} \\
\texttt{  self\_attn = SelfAttention(num\_heads, hidden\_dim)} \\
\texttt{  mlp\_1 = Dense(hidden\_dim, hidden\_dim * 2)} \\
\texttt{  mlp\_2 = Dense(hidden\_dim * 2, hidden\_dim)} \\
\texttt{  message\_head = Dense(hidden\_dim, message\_dim)} \\
\\
\textbf{IA Forward Pass (Message Generation):} \\
\texttt{obs\_features = activation(obs\_tok(partial\_obs))} \\
\texttt{context\_buf = concat([hidden\_state[1:], obs\_features[None]])} \\
\texttt{x = context\_buf + pos\_embed} \\
\texttt{y = self\_attn(ln1(x)) + x  \quad \textit{(residual connection)}} \\
\texttt{z = mlp\_2(activation(mlp\_1(ln2(y)))) + y  \quad \textit{(residual connection)}} \\
\texttt{message = tanh(message\_head(z[-1]))} \\
\texttt{return message, context\_buf} \\
\\ \hline \\
\textbf{Information Agent (IA) - World Model} \\
\texttt{$\vartriangleright$\hspace{0.5em}Additional prediction heads for dynamics modeling during training.} \\
\texttt{  next\_obs\_head = Dense(message\_dim + action\_dim, obs\_dim)} \\
\texttt{  reward\_head = Dense(message\_dim + action\_dim, 1)} \\
\texttt{  done\_head = Dense(message\_dim + action\_dim, 1)} \\
\texttt{  next\_msg\_head = Dense(message\_dim + action\_dim, message\_dim)} \\
\\
\textbf{IA World Model Forward Pass:} \\
\texttt{message, context\_buf = \textit{(as above)}} \\
\texttt{world\_input = concat([message, one\_hot(action, action\_dim)])} \\
\texttt{next\_obs = clip(next\_obs\_head(world\_input), -10.0, 10.0)} \\
\texttt{reward = clip(reward\_head(world\_input), -10.0, 10.0)} \\
\texttt{done = clip(sigmoid(done\_head(world\_input)), 1e-7, 1-1e-7)} \\
\texttt{next\_message = tanh(next\_msg\_head(world\_input))} \\
\\
\bottomrule[\heavyrulewidth]
\end{tabular}
\end{table*}

\begin{table*}[!htbp]
\centering

\caption{World Model benchmark (Integrated WM--Controller): shared Transformer encoder. \texttt{Dense(in, out)} denotes a fully connected layer. This table defines the shared encoder used by both the PPO controller and the auxiliary world model heads.}
\label{tab:world_model_benchmark_encoder}
\begin{tabular}{@{}p{0.95\linewidth}@{}}
\toprule[\heavyrulewidth]
\textbf{World Model Benchmark: Shared Transformer Encoder (context buffer)} \\
\midrule
\\
\textbf{Variables:} \\
\texttt{obs\_dim} = Flattened observation vector dimension (environment-dependent) \\
\texttt{hidden\_dim} = Transformer/feature dimension (default: 128) \\
\texttt{context\_len} = Transformer context buffer length \\
\texttt{num\_heads} = Self-attention heads \\
\texttt{activation} = \texttt{tanh} or \texttt{relu} (\texttt{tanh} by default) \\
\\ \hline \\
\textbf{Encoder Modules (single Transformer block)} \\
\texttt{$\vartriangleright$\hspace{0.5em}Fixed-length context buffer of encoded observations with learned positional embeddings.} \\
\texttt{  obs\_tok   = Dense(obs\_dim, hidden\_dim)} \\
\texttt{  pos\_embed = LearnedParameter(context\_len, hidden\_dim)} \\
\texttt{  ln1, ln2   = LayerNorm(), LayerNorm()} \\
\texttt{  self\_attn = SelfAttention(num\_heads, hidden\_dim)} \\
\texttt{  mlp\_1     = Dense(hidden\_dim, hidden\_dim * 2)} \\
\texttt{  mlp\_2     = Dense(hidden\_dim * 2, hidden\_dim)} \\
\\
\textbf{Encoder Forward Pass (produces features and updated context):} \\
\texttt{obs\_features = activation(obs\_tok(obs))} \\
\texttt{context\_buf  = concat([hidden\_state[1:], obs\_features[None]])} \\
\texttt{x = context\_buf + pos\_embed} \\
\texttt{y = self\_attn(ln1(x)) + x \quad \textit{(residual)}} \\
\texttt{z = mlp\_2(activation(mlp\_1(ln2(y)))) + y \quad \textit{(residual)}} \\
\texttt{features = z[-1]} \\
\texttt{return features, context\_buf} \\
\\
\bottomrule[\heavyrulewidth]
\end{tabular}
\end{table*}

\begin{table*}[!htbp]
\centering

\caption{World Model benchmark (Integrated WM--Controller): PPO actor--critic heads and one-step world model heads. Both consume the shared Transformer encoder features from Table~\ref{tab:world_model_benchmark_encoder}.}
\label{tab:world_model_benchmark_heads}
\begin{tabular}{@{}p{0.95\linewidth}@{}}
\toprule[\heavyrulewidth]
\textbf{World Model Benchmark: PPO Controller + One-step World Model Heads} \\
\midrule
\\
\textbf{Variables:} \\
\texttt{action\_dim} = Number of discrete actions (\texttt{env.action\_space(None).n}) \\
\texttt{obs\_dim} = Flattened observation vector dimension (environment-dependent) \\
\texttt{hidden\_dim} = Feature dimension (matches encoder) \\
\texttt{world\_model\_coef} = scalar coefficient for auxiliary world model losses \\
\\ \hline \\
\textbf{Controller (PPO Actor--Critic)} \\
\texttt{$\vartriangleright$\hspace{0.5em}Actor--critic heads for control.} \\
\texttt{  policy\_features = Dense(hidden\_dim, hidden\_dim)} \\
\texttt{  actor\_out       = Dense(hidden\_dim, action\_dim)} \\
\texttt{  critic\_out      = Dense(hidden\_dim, 1)} \\
\\
\textbf{Policy Forward Pass (mode=\texttt{'policy'}):} \\
\texttt{features, context\_buf = \textit{Shared Encoder Forward Pass (Table~\ref{tab:world_model_benchmark_encoder})}} \\
\texttt{pfeat  = activation(policy\_features(features))} \\
\texttt{logits = actor\_out(pfeat)} \\
\texttt{pi     = Categorical(logits)} \\
\texttt{value  = squeeze(critic\_out(pfeat))} \\
\texttt{return pi, value, context\_buf} \\
\\ \hline \\
\textbf{One-step World Model Heads (auxiliary)} \\
\texttt{$\vartriangleright$\hspace{0.5em}Predicts next observation, reward, and termination conditioned on (features, action).} \\
\texttt{  world\_model\_input = concat([features, one\_hot(action, action\_dim)])} \\
\texttt{  wm\_features        = Dense(hidden\_dim + action\_dim, hidden\_dim)} \\
\texttt{  next\_obs\_head     = Dense(hidden\_dim, obs\_dim)} \\
\texttt{  reward\_head        = Dense(hidden\_dim, 1)} \\
\texttt{  done\_head          = Dense(hidden\_dim, 1)} \\
\\
\textbf{World Model Forward Pass (mode=\texttt{'world\_model'}):} \\
\texttt{features, context\_buf = \textit{Shared Encoder Forward Pass (Table~\ref{tab:world_model_benchmark_encoder})}} \\
\texttt{wm\_in    = concat([features, one\_hot(action, action\_dim)])} \\
\texttt{wm       = activation(wm\_features(wm\_in))} \\
\texttt{next\_obs = clip(next\_obs\_head(wm), -10.0, 10.0)} \\
\texttt{reward   = clip(squeeze(reward\_head(wm)), -10.0, 10.0)} \\
\texttt{done     = clip(sigmoid(squeeze(done\_head(wm))), 1e-7, 1-1e-7)} \\
\texttt{return next\_obs, reward, done, context\_buf} \\
\\ \hline \\
\textbf{Training Usage (PPO + auxiliary WM losses)} \\
\texttt{$\vartriangleright$\hspace{0.5em}Rollouts are used for PPO losses and one-step prediction losses.} \\
\texttt{  L\_next\_obs = MSE(next\_obs\_pred, stop\_grad(next\_obs))} \\
\texttt{  L\_reward   = MSE(reward\_pred,  stop\_grad(clip(reward, -10, 10)))} \\
\texttt{  L\_done     = BCE(done\_pred,    stop\_grad(done))} \\
\texttt{  total\_loss = PPO\_loss + world\_model\_coef * (L\_next\_obs + L\_reward + L\_done)} \\
\\
\bottomrule[\heavyrulewidth]
\end{tabular}
\end{table*}

\section{Quantitative Analysis Details}
\subsection{Statistical Significance Testing}
\label{subsec:welch_t_test}
The performance metrics reported in Table~\ref{table: TTT} (Time-To-Threshold) and Table~\ref{table: MaxPerGeneralization} (Zero-Shot Performance) are means computed over 30 independent seeds which are subject to statistical variability. To determine whether the observed differences between CORAL and the baseline methods (PPO, World Model) are statistically meaningful, we performed pairwise hypothesis testing. For each environment and each pair of methods (e.g., CORAL vs. PPO), we used an independent two-sample \textbf{Welch's t-test}. The test does not assume that the variance of outcomes is equal across different methods. We report a difference as statistically significant if the test yields a p-value below our chosen significance level of $\alpha=0.05$. 

All \textbf{confidence intervals} in the paper are 95\% confidence intervals, calculated as $\bar{x} \pm 1.96 \cdot \frac{\sigma}{\sqrt{N}}$, where $\bar{x}$ is the sample mean, $\sigma$ is the sample standard deviation, and $N=30$ is the number of seeds.

\subsection{Time-to-Threshold (TTT) Calculation}

To provide a measure of sample efficiency that complements the learning curves, we use a Time-to-Threshold (TTT) analysis. TTT quantifies the number of environment steps required for an agent to reliably fulfill the task, or at least reach a high level of performance. Our calculation follows a two-pass process to ensure fair comparisons across environments with different reward scales and performance ceilings.

\begin{enumerate}
    \item \textbf{Performance Thresholds}: Initially, for each evaluation environment, we determine the maximum asymptotic performance achieved across all runs of all methods. This sets an empirical "best-case" performance for the task. The performance threshold for each environment is then set to 90\% of this maximum value. 
    \item \textbf{Calculating TTT and Success Rate}: In the second pass, for each individual seed of each method, we identify the first timestep at which the agent's episodic return meets or exceeds the calculated 90\% performance threshold. This timestep is recorded as the TTT for that run. If a run fails to reach the threshold within the maximum allowed timesteps for that environment, its TTT is considered undefined. The final TTT reported in Table~\ref{table: TTT} is the mean over all successful runs. To capture the reliability of each method, we also present the Success Rate (SR), which is the percentage of the 30 independent runs that successfully reached the performance threshold.
\end{enumerate}

\clearpage
\onecolumn

\section{Analysis of the Emergent Communicative Protocol during In-Context Adaptation}

Throughout the paper, we speculated that the emergent communicative protocol acts as an adaptive regularizer (one that exerts the most control when the CA is in greatest need of help resolving ambiguity). Here, we provide the analysis of the ICE metric from the main text over our entire testing family. Fig.~\ref{fig:ice} extends Fig.~\ref{fig:ice-compact} from the main text but with the whole test suite. Each row corresponds to a single unseen environment. The top row shows the learning curve, and the bottom row shows the corresponding ICE trajectory.

\begin{figure*}[!ht]
    \centering
    \includegraphics[width=.8\linewidth]{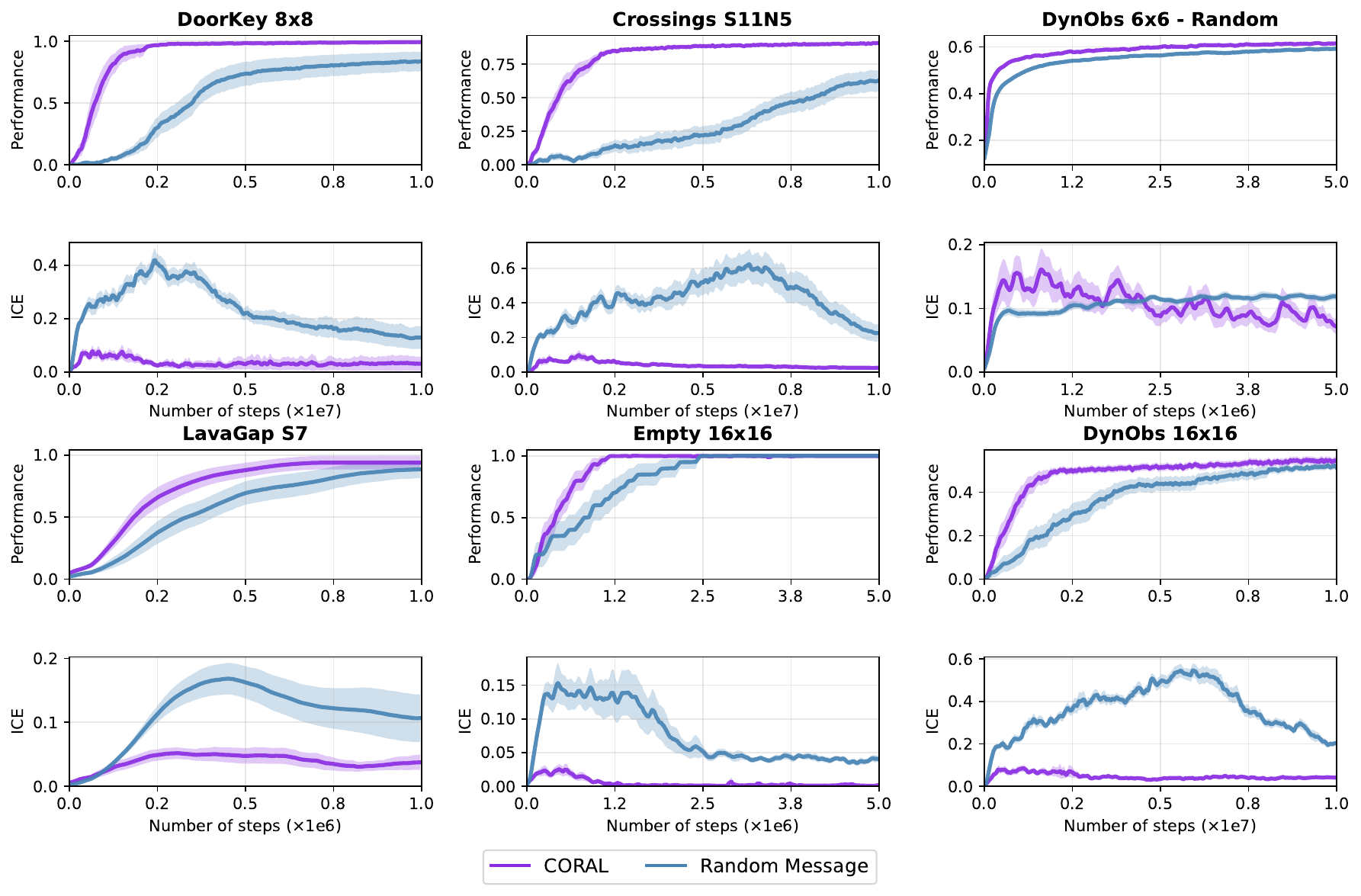}
    \caption{Analysis of Influence of Communication during In-Context Adaptation. Learning curves (top row) are paired with the corresponding Instantaneous Causal Effect metric (bottom row) for six unseen environments. ICE measures the per-step policy shift induced by communication (Eq.~\ref{eq:ice}). The \textcolor[HTML]{4682B4}{Random Message} baseline induces a high but unproductive ICE by distracting the agent. Conversely, the  ICE of the \textcolor[HTML]{923BE5}{CORAL} agent rises as the agent learns and then recedes as the policy converges.}
    \label{fig:ice}
    \vspace{-10pt}
\end{figure*}

\paragraph{Productive vs. Distractive Influence.}

One of the key takeaways from these plots is the difference between the magnitude of influence versus the usefulness of influence. Throughout training, the \textcolor[HTML]{4682B4}{Random Message baseline} maintains a very high ICE. In several environments, its ICE is greater than that of \textcolor[HTML]{923BE5}{CORAL}. But because this influence is not correlated with task reward, it fails to solve the task consistently in harder environments (e.g., DoorKey 8x8 and Crossings S11N5). This underscores that High ICE magnitude alone is insufficient for successful adaptation; dependence must be productive, driving the CA toward high-value regions of the policy space. Random messages have a strong and ``distractive'' influence because they induce large shifts in the policy distribution that are uninformative about the goals of the task.

\paragraph{Temporal Dynamics of Dependency.}
If we contrast the temporal dynamics of CORAL's lines with those of the Random Message baseline, we can draw two conclusions about when and how CORAL relies on communication:

\begin{itemize}

\item \textbf{Dependency arises during exploration: } For the first $0.2 \times 10^7$ steps, the ICE metric trends sharply upward. During this phase, we observe that the Control Agent's policy is becoming increasingly sensitized to the message context provided by the Information Agent. The CA begins to understand how $m_t$ correlates with reward, and this drives an increasing KL divergence between $\pi(\cdot|o,m)$ and $\pi(\cdot|o,0)$.

\item \textbf{Dependency fades during convergence: } During training where performance improves towards the asymptote (Empty 16x16, DoorKey 8x8), ICE gradually decreases. This would indicate a loss of informational dependency: once the CA has a good policy given the current episode's layout, then additional information from the message has less impact on the action distribution. Instead of driving behavior, the message serves more to confirm what the agent is already planning to do, ceding control back to the CA local policy.
\end{itemize}

\section{Ablation Studies}
\label{appendix:ablation}
\subsection{Importance of Message Coherence}

To validate the importance of using the Temporal Coherence Loss $\mathcal{L}_{\text{Coh}}$, we conducted an ablation study in the Information Agent's training. $\mathcal{L}_{\text{Coh}}$ is designed to promote temporal coherence in the emergent communication protocol, hypothesizing that a more predictable and consistent message stream brings a more stable learning signal for the Control Agent.

To test this, we pre-trained an \textcolor[HTML]{696969}{IA without the $\mathcal{L}_{\text{Coh}}$} to observe its ability to guide a new CA. The results presented in Table~\ref{fig:ablation_message_coh}
show that across all tested environments, the full CORAL agent consistently demonstrates superior sample efficiency and less variance compared to \textcolor[HTML]{696969}{its variant without $\mathcal{L}_{\text{Coh}}$}. The performance degradation is more noticeable in more complex tasks. 
\begin{figure*}[!ht]
    \centering
    \includegraphics[width=.9\linewidth]{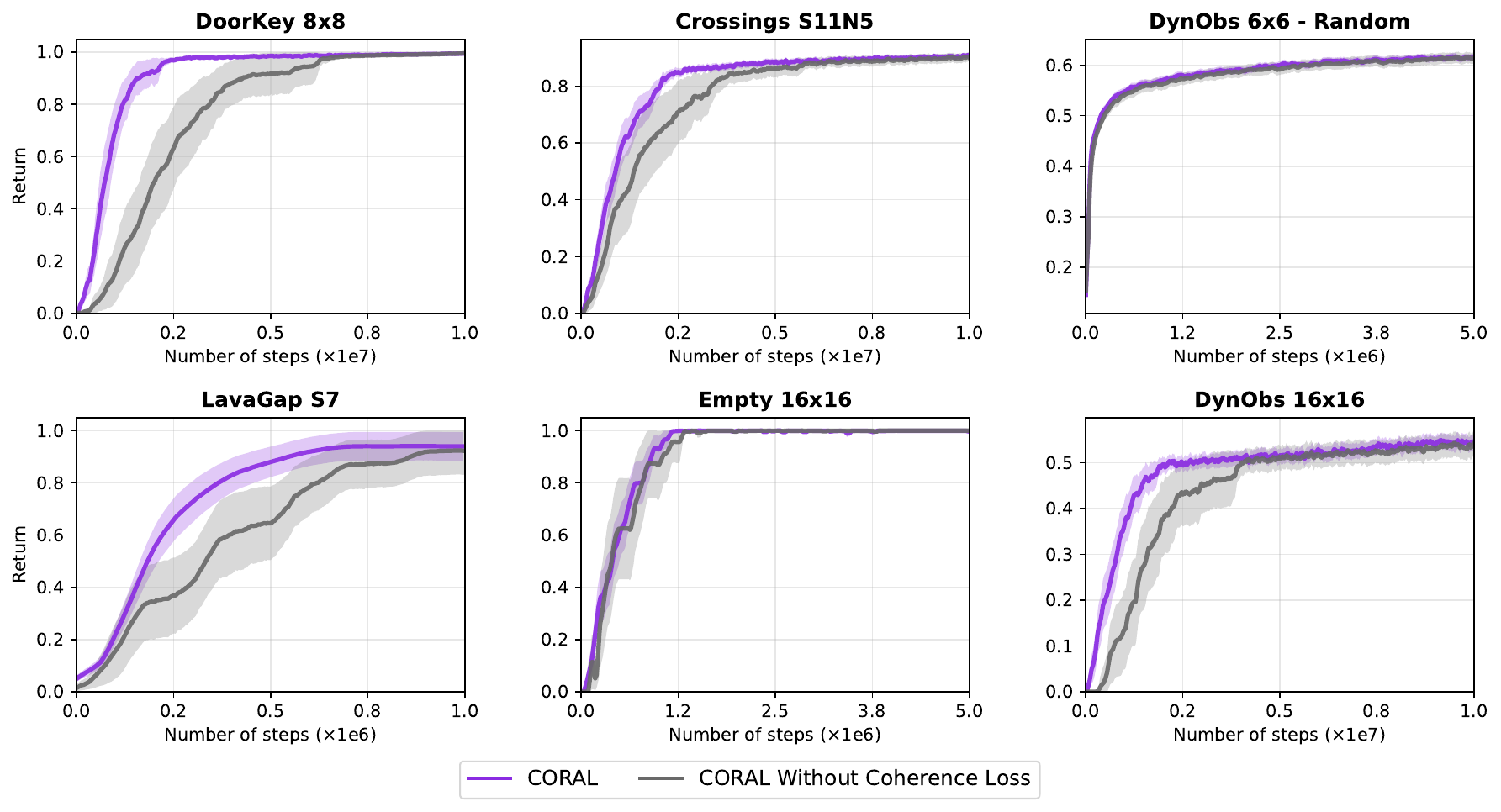}
    \caption{Ablation study on the Message Coherence Loss Learning curves show the mean episodic return ($\pm95\%$ confidence interval) across 30 multiple seeds for a randomly initialized Control Agent paired with a pre-trained, frozen CORAL IA. \textcolor[HTML]{923BE5}{CORAL} demonstrates better sample efficiency and asymptotic performance compared to an ablated version where the Information Agent was pre-trained \textcolor[HTML]{696969}{without the Temporal Coherence Objective}. The consistent performance degradation in the ablated agent across unseen environments demonstrates that promoting temporal coherence in the communication protocol is critical for achieving better sample efficiency.}
    \label{fig:ablation_message_coh}
\end{figure*}

\subsection{Importance of Causal Influence Objective}

\paragraph{Why Causal Influence Objective Matters.} 
\label{app:ablation_causal} 

In Fig.~\ref{appendix:nocausal}, we include the ablation study presented in the main text over the test suite. Results are plotted against CORAL without $\mathcal{L}_{\text{Causal}}$.

\paragraph{Passive vs. Pragmatic Communication.} 
We see that simply optimizing the world dynamics prediction $\mathcal{L}_{\text{Dyn}}$ and message coherence $\mathcal{L}_{\text{Coh}}$ cannot achieve the same level of fast adaptation. While agents trained with only these objectives do exhibit reasonable performance, our ablated agent takes almost twice as many interactions to match CORAL. This provides evidence that incentivizing the IA to send only information that changes the policy of the CA acts as a bottleneck on information flow, forcing the CA to learn and adapt to these messages right away instead of attempting to parse out relevant details from a pixel-level description of the state.

\paragraph{Asymptotic Performance Degrades.} 
Another observation we can make when looking at environments with more stochastic dynamics and rewards (\textit{LavaGap S7}, \textit{DynObs 6x6}) is that ablating $\mathcal{L}_{\text{Causal}}$ significantly degrades the final performance of the emergent protocol.

\begin{figure*}[!ht]
    \centering
    \includegraphics[width=0.85\linewidth]{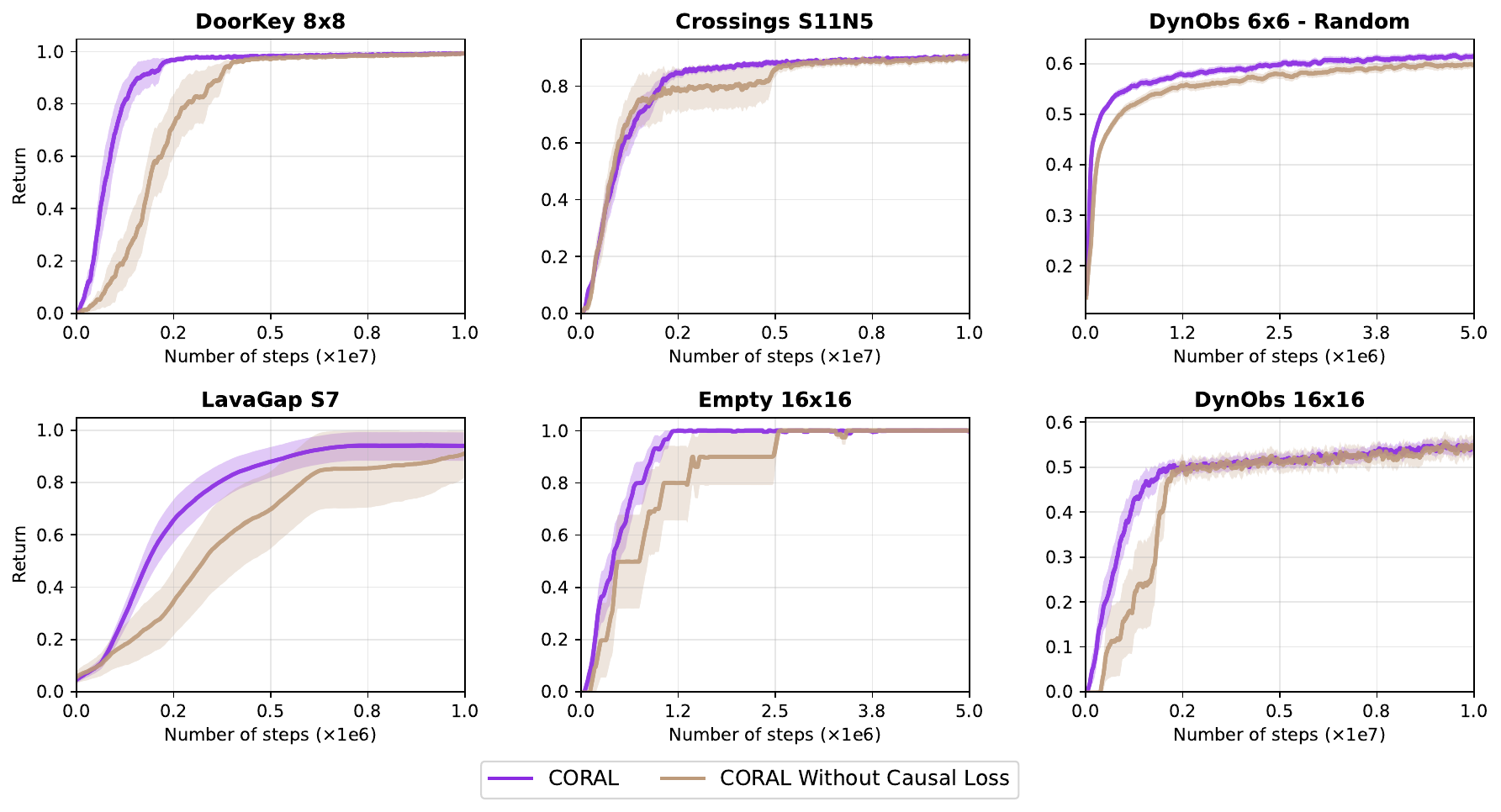}
    \caption{Ablation of Causal Influence Loss. Comparison between \textcolor[HTML]{8A2BE2}{CORAL} and an ablation \textcolor[HTML]{BD9A7A}{without causal loss} across unseen testing environments. While both methods use a pre-trained World Model, removing the Causal Influence objective degrades sample efficiency and reduces performance in stochastic tasks (\textit{LavaGap}, \textit{DynObs}), confirming the need to incentivize active influence for rapid adaptation.}
    \label{appendix:nocausal}
    \vspace{-5pt}
\end{figure*}

\newpage
\subsection{Information Agent Architecture}

Our Information Agent's design is based on a Transformer architecture to model the history of observations. We conducted an ablation study comparing it against a more traditional recurrent alternative. We implemented and pre-trained a version of the CORAL Information Agent where the Transformer block was replaced by a Gated Recurrent Unit (GRU) cell~\cite{chung2014empirical}. The GRU-based IA has comparable parameters and size, and is trained with the exact set of objectives as our main Transformer-based agent.

The in-context guided adaptation performance influenced by both architectures is presented in Table~\ref{fig:performance_GRU}. The resulting learning curves show that CORAL Transformer outperforms or matches CORAL GRU in all environments.

\begin{figure*}[!ht]
    \centering
    \includegraphics[width=.85\linewidth]{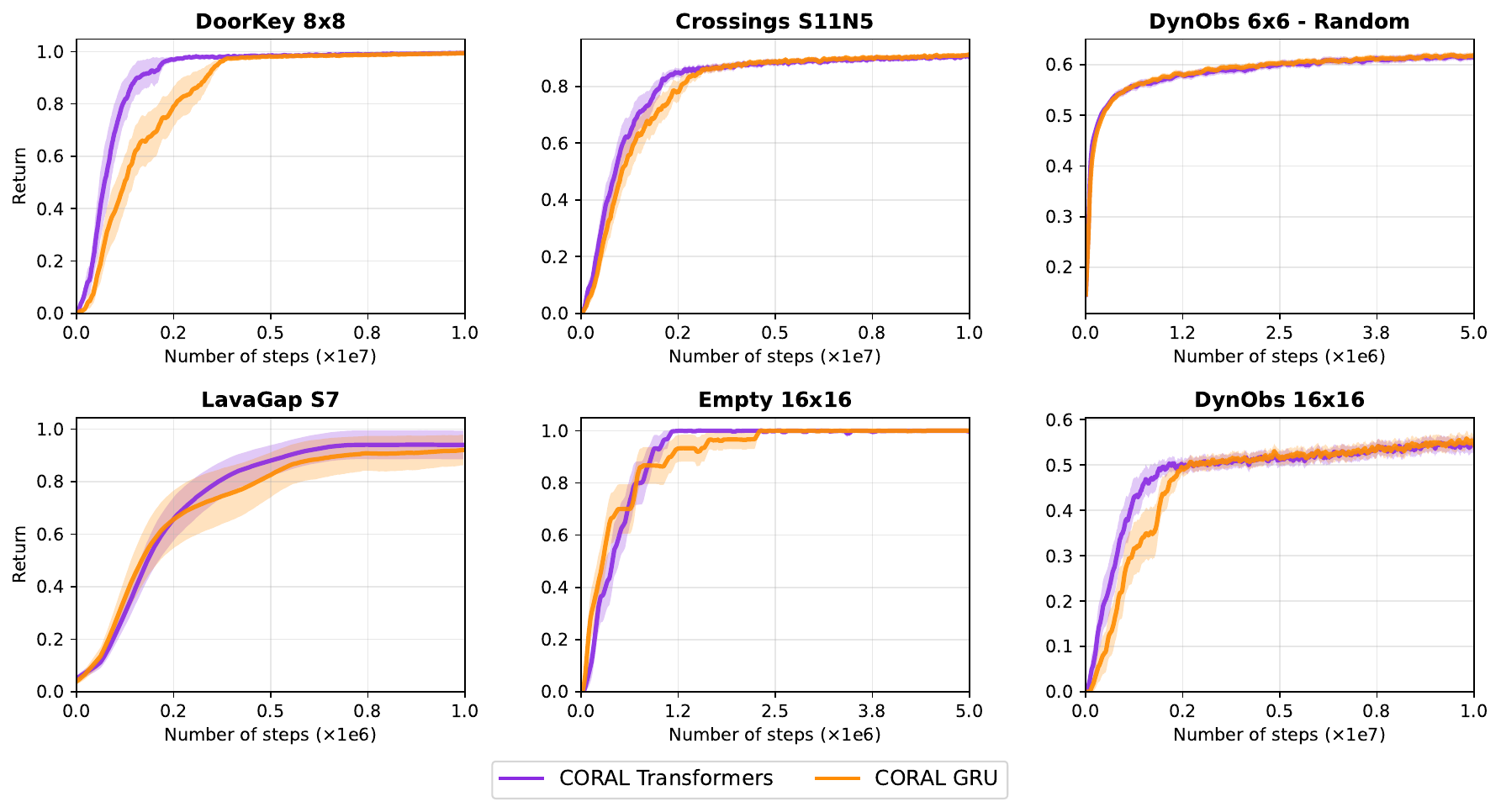}
    \caption{Ablation study on the Information Agent's architecture. Learning curves compare the in-context adaptation performance when the IA is implemented with a Transformer-based model versus a GRU-based alternative.}
    \label{fig:performance_GRU}
    \vspace{-5pt}
\end{figure*}

\newpage

\subsection{Message Dimensionality}

We performed an ablation study to analyze the sensitivity of CORAL to the dimensionality of the message vector. We pre-trained and deployed three versions of the framework using different message sizes: 16, 32, and 64, while keeping all other hyperparameters unchanged. 

Table~\ref{fig:message_dim} shows that CORAL is relatively robust to this choice, with all variants learning efficiently, especially compared to the baselines studied in our experimental results. \textbf{Message Size of 32} consistently achieves the best or near-best performance across the majority of tasks. The performance of the agent using a message size of 16 is often marginally slower, while the largest size of 64 also exhibits a minor lag in learning speeds in several environments. Empirically, a message size of 32 demonstrated the most consistent performance, justifying its use in our main experiments.

\begin{figure*}[!ht]
    \centering
    \includegraphics[width=.9\linewidth]{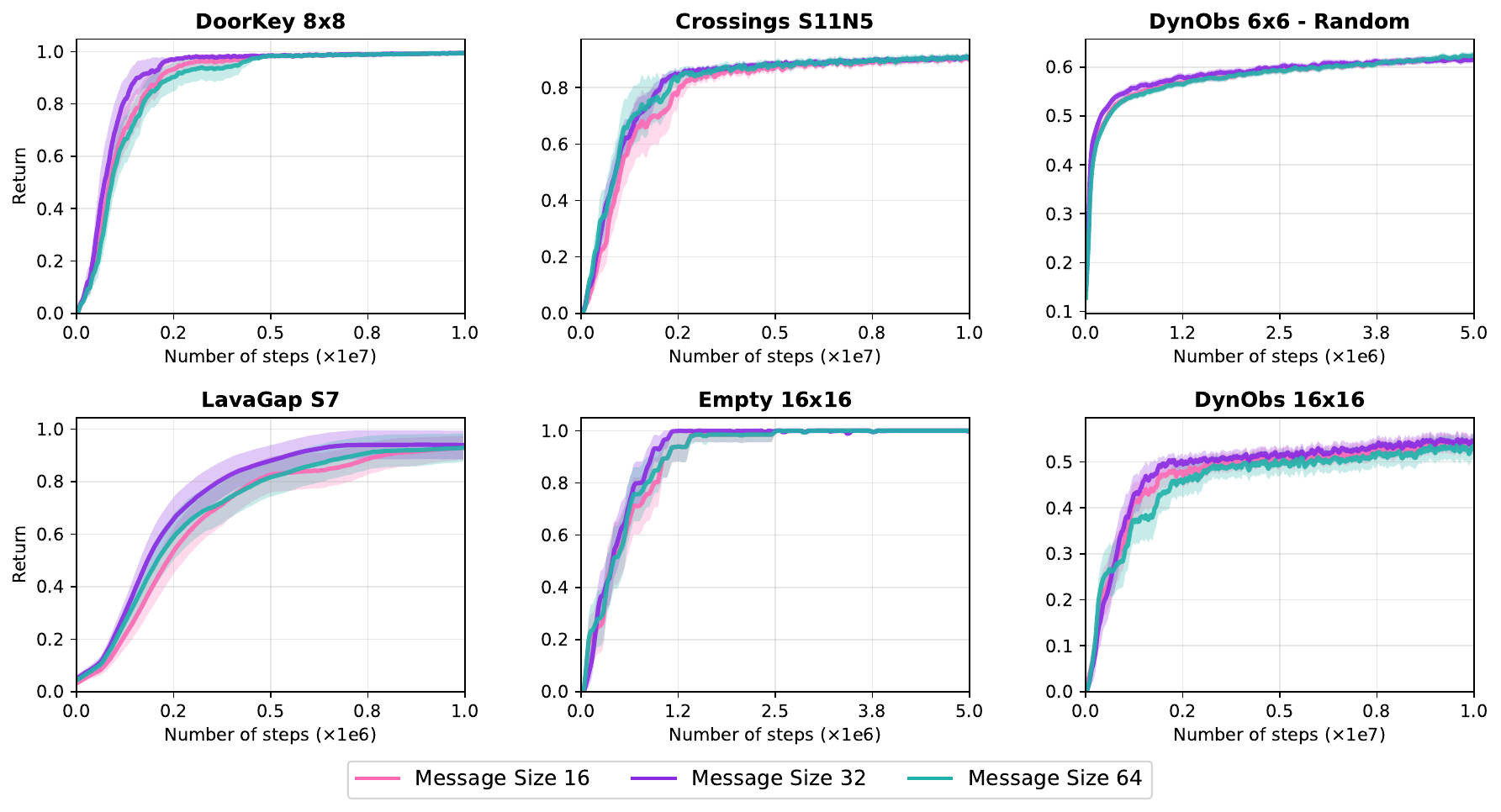}
    \caption{Ablation study on message dimensionality. In-context adaptation performance of CORAL with different message vector sizes (16, 32, 64). While CORAL is robust to this hyperparameter, a message of size 32 consistently provides a slight performance advantage.}
    \label{fig:message_dim}
\end{figure*}

\section{What Does the World Model In-Context Learn? A Gradient Dynamics Perspective}
\label{sec:transformer}

Recall that in the neural network structure for the information agent (IA) in CORAL, the embedded observations are processed through a series of self-attention and feed-forward layers, and the parameters of these networks are trained via a loss function that consists of prediction error for the next observation, temporal incoherence, and causal influence. More generally, the self-attention mechanism appears to be a vital component of in-context reinforcement learning, see e.g.,~\cite{moeini2025surveyincontextreinforcementlearning}. 
Therefore, a better understanding of this self-attention mechanism is beneficial not only to us but reinforcement learning community in general.

It is commonly accepted that a model for analyzing self-attention in-context learning by investigating the linear self-attention function, see e.g.,~\cite{wangtransformers}. In a basic setup, the prompts are represented by a $(d+1)\times (n+1)$ real matrix $Z$, two $(d+1)\times (d+1)$ parameter matrices, $P, Q$; and a $(n+1)\times (n+1)$ masking matrix, $M$, which is designed for in-context learning to reflect the specific data structure in the prompt where the last column of $Z$ is the query and the first $n$ columns as the context, see e.g.,~\cite{von2023transformers}. The linear attention function is defined as 
\begin{align*}
\Gamma(Z; P, Q) := PZM(Z^\top QZ),
\end{align*}
which is naturally a $(d+1)\times (n+1)$ matrix. 

In a network of $L$ self-attention layers with $P_\ell, Q_\ell$, $\ell=0,1,\ldots, L-1$, the forward pass takes the form of 
\begin{align*}
Z_{\ell+1} = Z_{\ell} -\frac{1}{n}\Gamma(Z_\ell; P_\ell, Q_\ell),
\end{align*}
with the following structural assumption,
\begin{align*}
Z_0=\begin{bmatrix} X\\ Y \end{bmatrix}, P_\ell=\begin{bmatrix} 0 \\ p_\ell^\top \end{bmatrix}, \quad Q_\ell=\begin{bmatrix} A_\ell & 0 \\ q_\ell^\top & 0\end{bmatrix}, 
\end{align*}
with $X$ representing the feature data and $Y$ the labels. $p_\ell$ being an $n+1$-dimensional vector, $q_\ell$ being an $n$-dimensional vector, and $A_\ell$ an $n\times n$ matrix, for all $\ell=0,1, \cdots, L-1$. In the IA, $X$ is represented by embedded tokens such as observation, and $Y$ represents the corresponding consequences of CA actions.  
This architecture has been widely used in the analysis of linear attention mechanisms, see, e.g.,~\cite{ahn2023transformers,wangtransformers,dingoptimality}, and demonstrates its versatility and power in capturing the basic dynamics of the self-attention mechanism. This architecture ensures that the predictor takes the form of a multilinear function. Moreover, in both ~\cite{ahn2023transformers} and~\cite{dingoptimality}, it is shown that the optimal solutions of the training process of the transformer with a quadratic loss function, similar to the one for IA, possess.
Furthermore, it has been demonstrated for various structures, see, e.g.,~\cite{ahn2023transformers,wangtransformers,dingoptimality}, that the above forward pass can be viewed as an implementation of the gradient descent update for different optimization problems, thus demonstrating the algorithmic power of the (linear) self-attention architecture. In addition, for properly selected parameters, the forward pass indeed implements the temporal difference method for reinforcement learning as demonstrated in~\cite{wangtransformers}. From this perspective, CORAL offers greater flexibility in training and interaction with the environment.

Theorem~\ref{thm:forward1} introduces a general form of such a result on the forward pass for in-context transformers with the usual sparsity assumption found in the literature. While this result is similar to that in~\cite{dingoptimality}, the proof methodology is more succinct, and the relationship between the structural properties of the parameter matrices and the result is better revealed and explained. The basic arguments in~\cite{ahn2023transformers,wangtransformers,dingoptimality} are first to conclude that the outcome of the forward pass is in an affine form. We observe that this result is mainly the combined result of the masking matrix $M$ and the last column of $Q$.

The predictor is $Z_L[d+1, n+1]$, i.e., the $(d+1, n+1)$ entry of the matrix $Z_L$, so we only need to focus on the he sequence $Z_\ell[d+1, n+1]$, $\ell=0,1,\ldots, L-1$. Because,
\begin{align*}
M=\begin{bmatrix} I_n & 0\\ 0 &0\end{bmatrix},
\end{align*}
$Z_\ell[d+1, n+1]= Z_0[d+1, n+1]+ \xi_\ell^\top x_{n+1}$, this can be easily proved by induction since the increment
\begin{align*}
\Gamma(Z_\ell; P_\ell, Q_\ell)[[d+1, n+1]&=\sum_{i=1}^{d+1} \sum_{j=1}^{n+1} \sum_{k=1}^{n+1}\sum_{q=1}^{d+1} \sum_{m=1}^{d+1}  P_\ell[d+1, i]Z_\ell[i,j]M[j,k]Z_\ell[\ell,k]Q_\ell[q,m] Z_\ell[m, n+1]\\=&
\sum_{i=1}^{d+1} \sum_{j=1}^{n} \sum_{q=1}^{d+1} \sum_{m=1}^{d}  P_\ell[d+1, i]Z_\ell[i,j]Z_\ell[q,j]Q_\ell[q,m] Z_\ell[m, n+1]
\end{align*}
which is linear with respect to the vector $Z[\cdot, n+1]$, because written in vector form, it becomes, 
\begin{align*}
\Gamma(Z_\ell; P_\ell, Q_\ell)[[d+1, n+1]&=
P_\ell[d+1, \cdot]Z_\ell Z_\ell^\top Q_\ell x_{n+1}
\end{align*}
and with no $Z_0[d+1, n+1]$.  
In fact, we can have the following lemma,
\begin{lem}
\label{lem:recursion}
Under the assumption on $P_\ell$, $Q_\ell$ and $M$, there exist a sequence of vectors $\xi_\ell, \ell=0,1, \ldots, L-1$, such that 
$Z_\ell[d+1, \cdot]= Z_0[d+1, \cdot]+ \xi_\ell^\top X$.
\end{lem}
\begin{proof}
This can be proved by induction. It also leads to the recursive formula of $\xi_\ell$.
\begin{align*}
\xi_{\ell+1} =\xi_\ell- \frac{1}{n}P_\ell[d+1, \cdot]\begin{pmatrix}{\hat Z}_\ell {\hat Z}_\ell^\top &  {\hat Z}_\ell Z_\ell[d+1;\cdot]^\top \\
Z_\ell[d+1;\cdot] {\hat Z}_\ell^\top &Z_\ell[d+1;\cdot]Z_\ell[d+1;\cdot]^\top\end{pmatrix} Q_\ell 
\end{align*}
where ${\hat Z}_\ell$ denotes the matrix formed by the first $d$ rows of $Z_\ell$. And by the assumption on $P_\ell$, we know that ${\hat Z}_\ell=X_0$ for all $\ell=0,1,\ldots, L-1$, and thus,
\begin{align*}
\xi_{\ell+1} =\xi_\ell- \frac{1}{n}P_\ell[d+1, \cdot]\begin{pmatrix} X_0 X_0^\top &  X_0 Z_\ell[d+1;\cdot]^\top \\
Z_\ell[d+1;\cdot] X_0^\top &Z_\ell[d+1;\cdot]Z_\ell[d+1;\cdot]^\top\end{pmatrix} Q_\ell 
\end{align*}
Therefore, 
\begin{align*}
\xi_{\ell+1} =\xi_\ell- \frac{1}{n}P_\ell[d+1, \cdot]\begin{pmatrix} X_0 X_0^\top &  X_0 (Z_0[d+1, \cdot]+ \xi_\ell^\top X)^\top \\
(Z_0[d+1, \cdot]+ \xi_\ell^\top X) X_0^\top &(Z_0[d+1, \cdot]+ \xi_\ell^\top X) (Z_0[d+1, \cdot]+ \xi_\ell^\top X)^\top\end{pmatrix} Q_\ell 
\end{align*}
\end{proof}
Equipped with this lemma, we can show that the forward pass is equivalent to a pre-conditioned gradient descent step for multi-objective optimization, as discussed in more detail in the optimization formulation and related methodologies, see e.g.,~\cite{Marler2004}. 
\begin{thm}
\label{thm:forward1}
Consider the $L$-layer transformer parameterized by $p_l,A_l=-\begin{bmatrix}
 \bar{A_l}\\a_l^{\top}
\end{bmatrix}$ where $b_l\in\mathbb{R}^{d+1},\bar{A}_l\in\mathbb{R}^{d\times d},a_l\in\mathbb{R}^{d}$ for $l\in [L]$. The forward pass of the transformer is equivalent to optimizing the multi-objective function $R_I: \mathbb{R}^{d }\rightarrow \mathbb{R}^{d+1}$, $i=1,2$. Specifically, let $y_{n+1}^{(l)}$ be the bottom-right entry of the $l$th layer output.
Then 
$y_{n+1}^{(l)} = 
\langle \xi_l ,
x_{n+1} \rangle 
$
where $w_l$ is iteratively defined as follows:  
$\xi_0 =0$ and 
\begin{align}
\xi_{l+1} ^{\top}  =& \xi_{l}^{\top}- \frac{1}{n} p_l^{\top}  \left[\nabla R_1(\xi_{l}) A_\ell + \nabla R_2(\xi_{l}) \begin{pmatrix}0 \\ q^\top_\ell\end{pmatrix}\right].
\end{align} 
with $R_1(\xi)=\begin{pmatrix} X_0X_0^\top \xi \\ (Z_0[d+1, \cdot]\xi+ \frac12\xi_\ell^\top X\xi\end{pmatrix}$ and $R_2(\xi)=\begin{pmatrix}Z_0[d+1, \cdot]\xi+ \frac12\xi_\ell^\top X\xi \\ Z_0[d+1, \cdot]Z_0[d+1, \cdot]^\top \xi +\frac12\xi_\ell^\top X\xi +\frac{1}{3}\xi\|\xi\|^2 \end{pmatrix}$.
\end{thm}
\begin{proof}
From the Lemma, we have,
\begin{align*}
\xi_{\ell+1} = &\xi_\ell- \frac{1}{n} p_l^{\top}\begin{pmatrix} X_0 X_0^\top  \\
(Z_0[d+1, \cdot]+ \xi_\ell^\top X) X_0^\top \end{pmatrix} A_\ell \\ &- \frac{1}{n} p_l^{\top}\begin{pmatrix}  0& X_0 (Z_0[d+1, \cdot]+ \xi_\ell^\top X)^\top \\
0& (Z_0[d+1, \cdot]+ \xi_\ell^\top X) (Z_0[d+1, \cdot]+ \xi_\ell^\top X)^\top\end{pmatrix} \begin{pmatrix}0 \\ q^\top_\ell\end{pmatrix}. 
\end{align*}
It is straightforward to verify the Theorem.
\end{proof}

\end{appendix}
\fi

\end{document}